\documentclass{article} % For LaTeX2e
\usepackage{arxiv_style}
\usepackage{times}
\usepackage{amsmath,amsfonts, mathtools,bm, stmaryrd, wrapfig, tcolorbox, enumitem}
\usepackage{algorithmic, algorithm}

\definecolor{cadmiumgreen}{rgb}{0.0, 0.42, 0.24}
\definecolor{amethyst}{rgb}{0.6, 0.4, 0.8}

\newcommand{\be}{\begin{equation}}
\newcommand{\ee}{\end{equation}}
\newcommand{\bea}{\begin{equation*}\begin{aligned}}
\newcommand{\eea}{\end{aligned}\end{equation*}}

\newcommand{\mc}{\mathcal}

\newcommand{\dd}{\mathrm{d}}
\DeclareMathOperator{\st}{s.t.}

\newcommand{\Let}{\coloneqq}

\newcommand{\X}{X}

\newcommand{\cL}{\mathcal{L}}

\newcommand{\bbE}{\mathbb{E}}

\usepackage{graphicx}
\usepackage[colorlinks]{hyperref}
\usepackage{url}
\usepackage{cleveref}
\usepackage{multirow}
\usepackage{booktabs}
\usepackage{caption}
\usepackage{subcaption}

\title{Bellman Optimal Stepsize Straightening of Flow-Matching Models}

\author{Bao Nguyen\\
VinUniversity\\\texttt{bao.nn2@vinuni.edu.vn} \\
\And
Binh Nguyen \\
National University of Singapore\\\texttt{binhnt@nus.edu.sg} \\
\AND
Viet Anh Nguyen \\
Chinese University of Hong Kong\\\texttt{nguyen@se.cuhk.edu.hk}
}

\iclrfinalcopy % Uncomment for camera-ready version, but NOT for submission.

\begin{document}
\maketitle
\begin{abstract}
    Flow matching is a powerful framework for generating high-quality samples in various applications, especially image synthesis. However, the intensive computational demands of these models, especially during the finetuning process and sampling processes, pose significant challenges for low-resource scenarios. This paper introduces Bellman Optimal Stepsize Straightening (BOSS) technique for distilling flow-matching generative models: it aims specifically for a few-step efficient image sampling while adhering to a computational budget constraint. First, this technique involves a dynamic programming algorithm that optimizes the stepsizes of the pretrained network. Then, it refines the velocity network to match the optimal step sizes, aiming to straighten the generation paths. Extensive experimental evaluations across image generation tasks demonstrate the efficacy of BOSS in terms of both resource utilization and image quality. Our results reveal that BOSS achieves substantial gains in efficiency while maintaining competitive sample quality, effectively bridging the gap between low-resource constraints and the demanding requirements of flow-matching generative models. Our paper also fortifies the responsible development of artificial intelligence, offering a more sustainable generative model that reduces computational costs and environmental footprints. Our code can be found at \url{https://github.com/nguyenngocbaocmt02/BOSS}.
\end{abstract}

\section{Introduction}
\label{sec:intro}

There have been impressive advancements in deep generative models in recent
years, which constitute an appealing set of approaches capable of approximating
data distributions and generating high-quality samples, as showcased in
influential works such as
\cite{ref:ramesh2022hierarchical,ref:saharia2022photorealistic,ref:rombach2022high}. They are primarily driven by a category of time-dependent generative models
that utilize a predefined probability path, denoted as $\{\pi_t\}_{t\in[0, 1]}$.
This probability path is a process that interpolates between the initial noise distribution $\pi_0$ and the target data distribution $\pi_1$.
The training for these models can be broadly characterized as a regression task
involving a neural network function $v_\theta$ and a target ideal velocity $v_t(x)$:
\begin{equation*}
  \cL(\theta) \Let \bbE_{t \in [0, 1],~X_t \sim \pi_t}[\ell(v_\theta(X_t, t), v_t(X_t))].
\end{equation*}
Here, the velocity network $v_\theta$ maps any input data
$x$ at time $t \in [0, 1)$ to a vector-valued velocity quantity $v_\theta(x, t)$ and plays a crucial role in the generation of samples from the interpolation process through the relationship:
\begin{equation*}
    X_1 = X_0 + \int_{0}^{1} v_\theta(X_t, t) \dd t,
\end{equation*}
which is the solution of the ODE 
$\mathrm{d}\X_t = v_\theta(x_t, t)\mathrm{d}t$ with the boundary condition $X_{t=0}=X_0$.
A noteworthy class of algorithms that fits within this framework includes
denoising diffusion models
\citep{ref:ho2020denoising,ref:sohl2015deep,ref:song2020score} and the
more recent flow-matching/rectified-flow models
\citep{ref:liu2022flow,ref:lipman2022flow,ref:albergo2022building,ref:neklyudov2023action}.

The latter type of model extends the principles employed in training diffusion models to simulation-free continuous normalizing flows (CNF, \citealt{ref:chen2018neural}).
It is particularly attractive because it fixes the suboptimal alignment between noises and images of diffusion models by introducing a straight trajectory formula connecting them. This leads to (empirically observed) faster training and inference time than diffusion models. The rectified flow framework \cite{ref:liu2022flow} also includes a technique called \emph{reflow}, which gradually rectifies the probability paths. It significantly reduces the number of function evaluations needed for sampling and thus belongs to the family of distillation methods.

However, the standard reflow technique proposed in \citet{ref:liu2022flow} requires a significant amount of computational budget: on a small-dimensional dataset such as CIFAR-10 $(32\times32$ pixel images), it takes at least an additional 300,000 \textit{re}training iterations of the pretrained velocity networks to reach FID (Fr\'echet Inception Distance, \citealt{ref:heusel2017gans}) of 4.85 for 1-step generation. The additional retraining time can reach approximately 200 days of A100 GPU for distilling models for 1-step sampling on higher-dimensional scale datasets to achieve
competitive FID, as stated in an extension of the rectified flow framework \citep{ref:liu2023instaflow}. Motivated to fix this problem, in this work, we aim to distill the Rectified Flow model while satisfying the following objective.

\begin{tcolorbox}[colback=white!5!white,colframe=black!75!black, top=5pt,bottom=5pt]
Given a pretrained flow-matching velocity $v_\theta$ and a target of
$K$ number of function evaluations (NFEs), how can we adapt $v_\theta$ with a
proper sampling schedule to generate high-fidelity images using only a modest computational resource?
\end{tcolorbox}

\textbf{Contributions.} We propose BOSS, the Bellman Optimal Stepsize Straightening method, to finetune pretrained flow-matching models. Our proposal includes two phases. The first phase seeks the optimal $K$-element sequence $\Delta^*$ for the initial model $v_\theta$. The second phase utilizes $\Delta^*$ to retrain $v_\theta$ such that the retrained model $v_{\theta^*}$ performs better. With the proposed procedure, we straighten the velocity network with just about 10,000 retraining iterations while outperforming the standard reflow strategies regarding image quality.
Quantitatively, our procedure consistently achieves lower FID in unconditional image generation with four different datasets. Furthermore, as the additional results in Appendix~\ref{sec:lora} show, the straightening procedure using Low-Rank Adaptation (LoRA) can finetune only $2\%$ of the model's parameters, yet it performs competitively to that of full-rank finetuning.

\textbf{Related works.} Our approach is directly related to existing works on improving the sampling efficiency of diffusion and flow-matching models with \emph{training-based algorithms}. In \cite{ref:salimans2021progressive}, the authors proposed an approach to enhance the sampling speed of unguided diffusion models through iterative distillation. This is extended to the case of classifier-free guided diffusion models in \cite{ref:meng2023distillation}. In \cite{wang2023learning}, the authors propose a method leveraging reinforcement learning to automatically search for an optimal sampling schedule for Diffusion Probabilistic Models (DPMs), addressing limitations in hand-crafted schedules and the assumption of uniformity across instances. Our work has a few common features and motivations with \citet{watson2021learning}, which achieved significant speed-ups through dynamic programming and decomposed loss terms. However, their focus on individual Kullback–Leibler divergence loss that neglects the cumulative information loss during sampling. This results in images with reduced overall quality. In contrast, we focus on minimizing the local truncation error during the sampling procedure, which improves the image quality consistently across all budgets of NFEs. Moreover, we propose a finetuning method that allows faster sampling with just a few NFEs. Recent work by \cite{ref:song23consistency} introduced a framework that learns a model capable of mapping any point at any time to the trajectory’s starting point, called the \emph{consistency model}. After submitting this work, we discovered a concurrent study by \cite{li2023autodiffusion}. This work pointed out that using uniform stepsizes is suboptimal for diffusion model sampling and instead used evolutionary algorithms to search for the optimal stepsizes and score network architectures, with the FID score being the optimized metric. 

Within the context of Rectifed Flow/Flow Matching, \cite{ref:liu2022flow} proposed a reflow method that uses retraining to straighten the probability sampling path. This results in a low NFE sampling with favorable image quality. The recent work of \cite{ref:liu2023instaflow} takes this framework to a larger scale, demonstrating impressive results on high-resolution image datasets. However, both rely on computational intensive retraining procedures, which we improve in our work.

The other direction that aims to accelerate the sampling process of diffusion/flow matching models is \emph{training-free samplers} \citep{ref:song2020denoising,ref:bao2022analytic,ref:liu2022pseudo,ref:tachibana2021quasi,ref:zhang2022fast,ref:karras2022elucidating,ref:lu2022dpm,ref:zheng2023dpm}. Although required no additional training step, these works mainly relied on the properties of the SDE/probability flow ODE dynamics to propose heuristic solvers/diffusion noise schedulers. Therefore, verifying whether the proposed sampling stepsizes are optimal is hard.
\vspace{-0.5mm}
\section{Background}
\label{sec:state}
Suppose we are given a (pretrained) model $v_\theta(X_t, t)$ with parameter $\theta$, which is an estimator of the function $v$ from an ordinary differentiable equation (ODE) on the span $t \in [0, 1]$:
\be \label{eq:ode}
\dd X_t = v(X_t, t) \dd t.
\ee
In generative modeling with diffusion/flow matching models, this dynamic system is called the \emph{probability flow ODE} \citep{ref:song2020score,ref:lipman2022flow}. The estimator $v_\theta$ allows us to flow from the distribution $\pi_0$ (noises) to the distribution $\pi_1$ (real images) via the equation:
\begin{equation}
  \label{eq:rand-map}
  X_1 = X_0 + \int_{0}^{1} v_\theta(X_t, t) \dd t,
\end{equation}
where $X_0 \sim \pi_0$ and $X_1 \sim \pi_1$. 
In the context of our problem, $X_0$ is observable. $X_1$ is only determined by the equation~\eqref{eq:rand-map} which is a deterministic process that for each $X_0 = x_0$, there is only value $X_1 = x_1$ coupling with it through the following equation:
\[
  x_1 = x_0 + \int_{0}^{1} v_\theta(x_t, t) \dd t.
\]
We are interested in the low-cost estimate of the integral $\int_{0}^{1} v_\theta(x_t, t) \dd t$ with respect to $t$ over the interval $[0, 1]$. It is an essential concern when calculating the velocity field is computationally expensive. The amount of times calling $v_\theta$ is defined as the number of function evaluations (NFE).  

\textbf{First Order Sampling scheme.} To solve for the integration that appears in the sampling equation~\eqref{eq:rand-map}, it is necessary to invoke a numerical integrator that uses discretized time steps.  Any numerical integration scheme will induce truncation errors, which can be quantified in two forms: first, when we have the value at the previous time step $X_{\tau-\delta}$, the solver estimates the subsequent true value $X_{\tau}$ as $\hat X_{\tau}$, causing a \emph{local truncation error} $X_{\tau} - \hat X_{\tau}$. These local errors accumulate over the number of intervals, eventually resulting in a cumulative error known as the \emph{global truncation error}. The most popular discretization scheme is perhaps Euler's method: given a budget $K$ number of function evaluations (NFEs), the Euler uniform sampling computes the interval $\Delta = 1/K$, and the sample successively
\be \label{eq:euler-u}
x_{k/K}^i = x_{(k-1)/K}^i + v_\theta(x_{(k-1)/K}^i, (k-1)\Delta) \times \Delta \quad \forall k = 1, \ldots, K,
\ee 
with the initial condition $x_0^i \sim \pi_0$. We denote this uniform sampling scheme by $\mc E^U(K)$. Euler's method with uniform stepsizes $\Delta$ has local truncation error $O(\Delta^2)$, and global truncation error $O(\Delta)$.
One can generalize the Euler sampling with \textit{non-uniform} intervals by dividing the time domain $[0, 1]$ into unequal intervals with timestamps $0 = \tau_0 < \tau_1 \ldots < \tau_K = 1$, and sample successively
\begin{equation} \label{eq:euler-nonu}
  x_{\tau_k}^i = x_{\tau_{k-1}}^i + v_\theta(x_{\tau_{k-1}}^i, \tau_{k-1}) \times (\tau_k - \tau_{k-1}) \quad \forall k = 1, \ldots, K,
\end{equation}
with the initial condition $x_0^i \sim \pi_0$. The timestamps equivalently determine the stepsize $\tau_k - \tau_{k-1}$ for each sampling iteration. We denote this scheme by $\mc E(\{\tau_0, \tau_1, \ldots, \tau_K\})$. If the timestamps $\tau_k$ are equally spaced in $[0, 1]$, then we obtain the equivalence $\mc E(\{\tau_0, \tau_1, \ldots, \tau_K\})\equiv \mc E^U(K)$. Since the reflow procedure in \cite{ref:liu2022flow} deals exclusively with Euler's method for being the fastest with a fixed computational budget, we focus only on this method in our paper.

\section{Optimal Sampling Stepsizes} \label{sec:step}

The objective presented in \Cref{sec:intro} can be defined more rigorously as follows.
Given a pretrained model $v_\theta$ and an insignificant value of $K$, we aim to find the optimal value $\theta^*$ and sequence $\Delta^*$ such that $\mc E(.,\Delta^*)$ is a reasonable estimate for the coupling sample $x_1$ of any $x_0$. This can be posed as an integer optimization problem of finding the best schedule for sampling. Given a fixed budget of $K$ NFEs, we find a schedule $\{\tau_0, \tau_1, \ldots, \tau_K\}$ satisfying $0 = \tau_0 < \tau_1 < \ldots < \tau_K = 1$ and that the associated Euler non-uniform sampling scheme $\mc E(\{\tau_0, \tau_1, \ldots, \tau_K\})$ has minimal sampling error for the pretrained velocity $v_\theta$. In Section~\ref{sec:cost}, we describe our estimate of the sampling error for any valid schedule. Section~\ref{sec:integer} presents an integer programming formulation to find the optimal stepsizes for sampling, and Section~\ref{sec:bellman} provides a dynamic programming algorithm to find the optimal schedule.

\subsection{Sampling Error Estimation} \label{sec:cost}
Given any two arbitrary timestamps $0 \le t_j < t_{k} \le 1$, we are interested in estimating the \textit{local} Euler truncation error, i.e., measuring the discrepancy between the true value
\[
X_{t_{k}} = X_{t_j} + \int_{t_j}^{t_{k}} v_\theta(X_t, t) \dd t
\]
and the one-step Euler sampling value
\[
X_{t_{k}}^\mc E = X_{t_j} + v_{\theta}(X_{t_j}, t_j) \times (t_{k} - t_j),
\]
where $X_{t_j}$ is sampled from the distribution $\pi_{t_j}$ which is induced by the initial distribution $\pi_0$ of $X_0$ and the ODE~\eqref{eq:ode}. This local truncation error can be formalized as
\[
c_{jk}^{\mathrm{truncation}} \Let \mathbb{E}_{X_{t_j} \sim \pi_{t_j}} \Big[\Big\| \int_{t_j}^{t_{k}} v_\theta(X_t, t) \dd t -  v_{\theta}(X_{t_{j}}, t_{j}) (t_{k} - t_{j})  \Big\|^2_2 \Big].
\]
Unfortunately, computing $c_{jk}^{\mathrm{truncation}}$ is computationally intensive because of both the expectation operator and the integration. We instead employ the following two simplifications:
\begin{enumerate}[leftmargin = 5mm]
    \item We fix the possible choice of time-stamps: for a sufficiently large number $K^{\max}$, the anchoring timestamps are $\{t_k\}_{k=0, \ldots, K^{\max}}$ with $t_k = k/K^{\max}$.  In doing so, we have restrained the space of all possible sampling schedules to the combinations of finite anchoring timestamps $\{t_k\}$. Later in the numerical experiments, we choose $K^{\max} = 100$, leading to the anchoring timestamps $\{0, 0.01, 0.02, \ldots, 0.99, 1\}$.
    \item We approximate the local truncation error $c_{jk}^{\mathrm{truncation}}$ for any two anchoring timestamps $t_j < t_k$ by a sample average estimator $c_{jk}$ that is constructed as follows:
    \begin{equation}
    \label{eq: local_truncation_error}
    c_{jk} = \frac{1}{N} \sum_{i=1}^N c_{jk}^i, \quad \mathrm{where} \quad c_{jk}^i = \|x^i_{t_k} - x^i_{t_j} - v_{\theta}(x^i_{t_j}, t_j) \times (t_k - t_j) \|^2_2,
    \end{equation}
    where for sample $i$, the noise $x_0^i$ is drawn from $\pi_0$, and $x^i_{t_j}$ and $x^i_{t_k}$ are extracted from Euler uniform sampling path starting from $x_0^i$, taken at time $t_j$ and $t_k$, respectively.
\end{enumerate}

\begin{wrapfigure}{h}{0.5\textwidth}
    \vspace{-11mm}
    \begin{minipage}{1\linewidth}
    \begin{figure}[H]
        \includegraphics[width=1\linewidth]{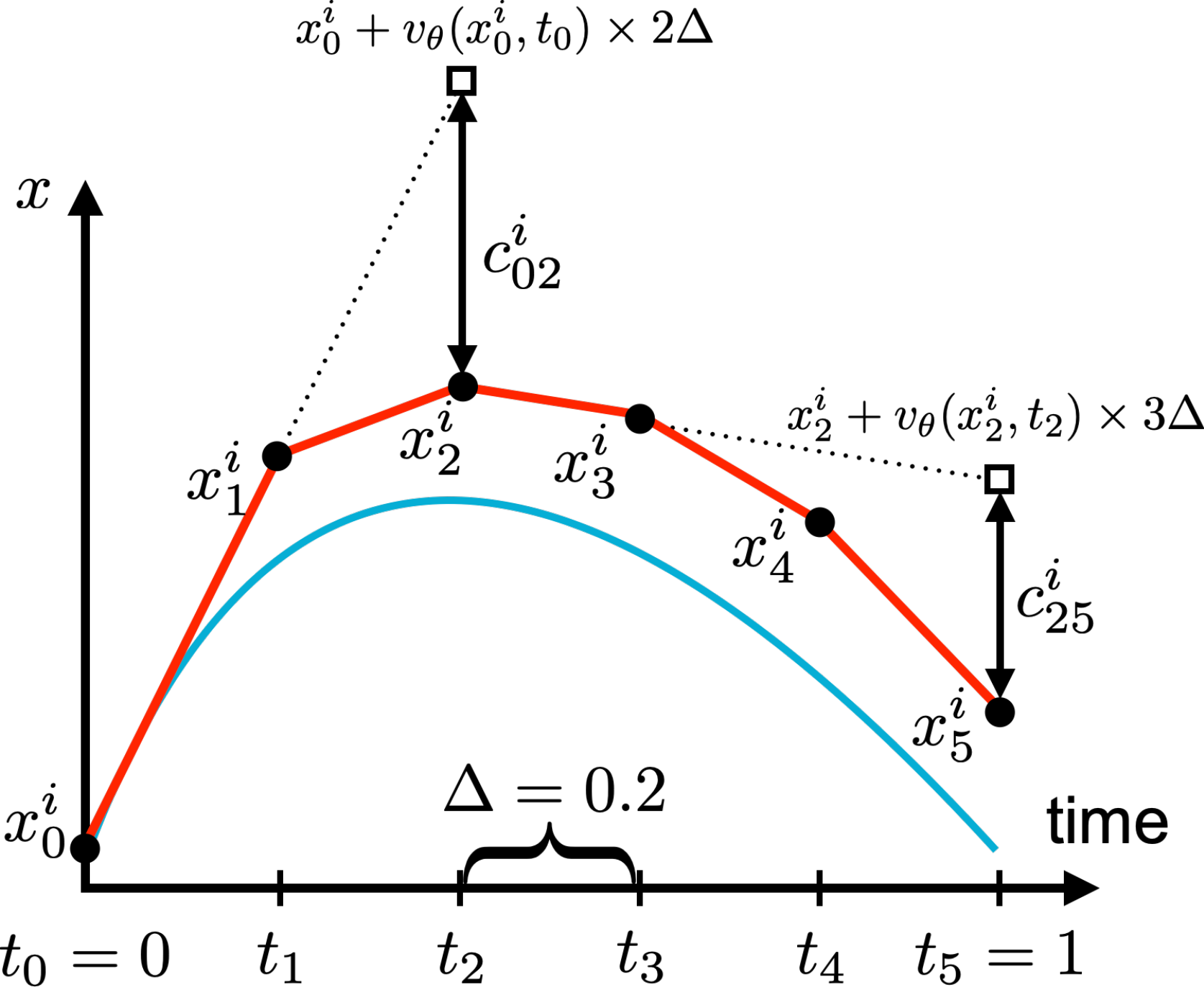}
        \caption{An example with $K^{\max} = 5$ to illustrate the computation of the sampling error.}
        \label{fig:cost}
    \end{figure}
    \end{minipage}
\end{wrapfigure}

Figure~\ref{fig:cost} illustrates how the sampling error $c_{jk}^i$ are calculated for the simple case with $K^{\max} = 5$ (or equivalently, with a uniform interval $\Delta = 0.2$). First, noise $x_0^i$ is drawn from $\pi_0$, and the blue curve depicts the nonlinear trajectory following the ODE~\eqref{eq:ode}. The piecewise linear trajectory is the path generated by the uniform Euler sampling with $K^{\max}$ NFEs, leading to the observed trajectory $\{x_k^i\}_{k=0, \ldots, K^{\max}}$. For a concrete example of computing $c_{25}^i$, we measure the difference between the value of a one-step Euler sampling from $t_2$ to $t_5$ with a stepsize $t_5 - t_2 = 3 \Delta$ to obtain $x_2^i + v_{\theta}(x_2^i, t_2) \times 3 \Delta$, and the observed value $x_5^i$. Intuitively, we can view $c_{jk}$ as the difference between the Euler one-step and the Euler $(k-j)$-step uniform sampling between $t_j$ and $t_k$.
\clearpage
One may observe that $c_{jk}$ is only an approximation of the true local truncation error $c_{jk}^{\mathrm{truncation}}$ because $c_{jk}$ is computed based on the Euler trajectory (red piecewise linear path in Figure~\ref{fig:cost}), while the truncation error $c_{jk}^{\mathrm{truncation}}$ should be computed based on the nonlinear trajectory of the ODE (blue curve in Figure~\ref{fig:cost}).
One downside of using $c_{jk}$ is that for any two consecutive timestamps $t_j$ and $t_{j+1}$, we have $c_{j,(j+1)} = 0$. This downside can be mitigated by taking $K^{\max}$ sufficiently large. On the other hand, as $K^{\max}$ gets large, calculating all the values $c_{jk}$ is computationally intensive because there are, in total, $K^{\max}(K^{\max}-1)/2$ pair of timestamps whose errors are to be computed. Nevertheless, we demonstrate empirically in \Cref{sec:experiments} that even when $c_{jk}$ is computed using a small number $N$ of samples, the resulting optimal schedule already demonstrates a superior performance vis-\`{a}-vis competing methods.

\subsection{Integer Programming Formulation} \label{sec:integer}

As we now describe, finding the optimal sampling schedule can be formulated as a network-flow-based problem~\citep{ref:ahuja1993network}. First, construct a graph of $K^{\max}+1$ nodes; each node represents one timestamp, see Figure~\ref{fig:integer}. There is an edge connecting node $t_j$ to node $t_k$ if $t_j < t_k$, and this edge is associated with a sampling error cost $c_{jk}$, computed in Section~\ref{sec:cost}. For a target of $K$ NFEs, the optimal sampling schedule is a path that traverses from the source node $t_0$ to the sink node $t_{K^{\max}}$ that is comprised of exactly $K$ edges. This path can be recovered from the optimal solution of the problem
\begin{equation} \label{eq:prob}
    \begin{array}{cll}
        \min & \displaystyle \sum_{j=0}^{K^{\text{max}}-1} \sum_{k=j+1}^{K^{\text{max}}} c_{jk} z_{jk} \\ [1.5ex]
        \st & \sum_{j=0}^{K^{\text{max}}-1} \sum_{k=j+1}^{K^{\text{max}}} z_{jk} = K \\ [1.5ex]
        & \sum_{k=1}^{K^{\text{max}}} z_{0k} = 1, \quad \sum_{j=0}^{K^{\text{max}}-1} z_{jK^{\text{max}}} = 1 \\ [1.5ex]
     &\sum_{k=0}^{j-1} z_{kj} = \sum_{k=j+1}^{K^{\text{max}}} z_{jk} &\forall j \in  \llbracket 1,  K^{\text{max}-1}\rrbracket \\
     & z_{jk} \in \{0, 1\} & \forall 0 \le j < k \le  K^{\max}.
    \end{array}
\end{equation}
Above, $z_{jk} \in \{0, 1\}$ is a binary decision variable, $z_{jk} = 1$ if the path takes a one-step sampling from time $t_j$ to time $t_k$. The objective function of~\eqref{eq:prob} minimizes the path's accumulated sampling error, which approximates the \textit{global} truncation error of the Euler sampling with the corresponding step sizes. The first constraint indicates that the path should consist of exactly $K$ edges; the second constraint imposes that $t_0$ and $t_{K^{\max}}$ are the source and sink nodes, respectively. Finally, the last set of constraints is the flow conservation on each intermediary node  between $t_0$ and $t_{K^{\max}}$. 
    \begin{figure}
        \centering
        \includegraphics[width=0.8\linewidth]{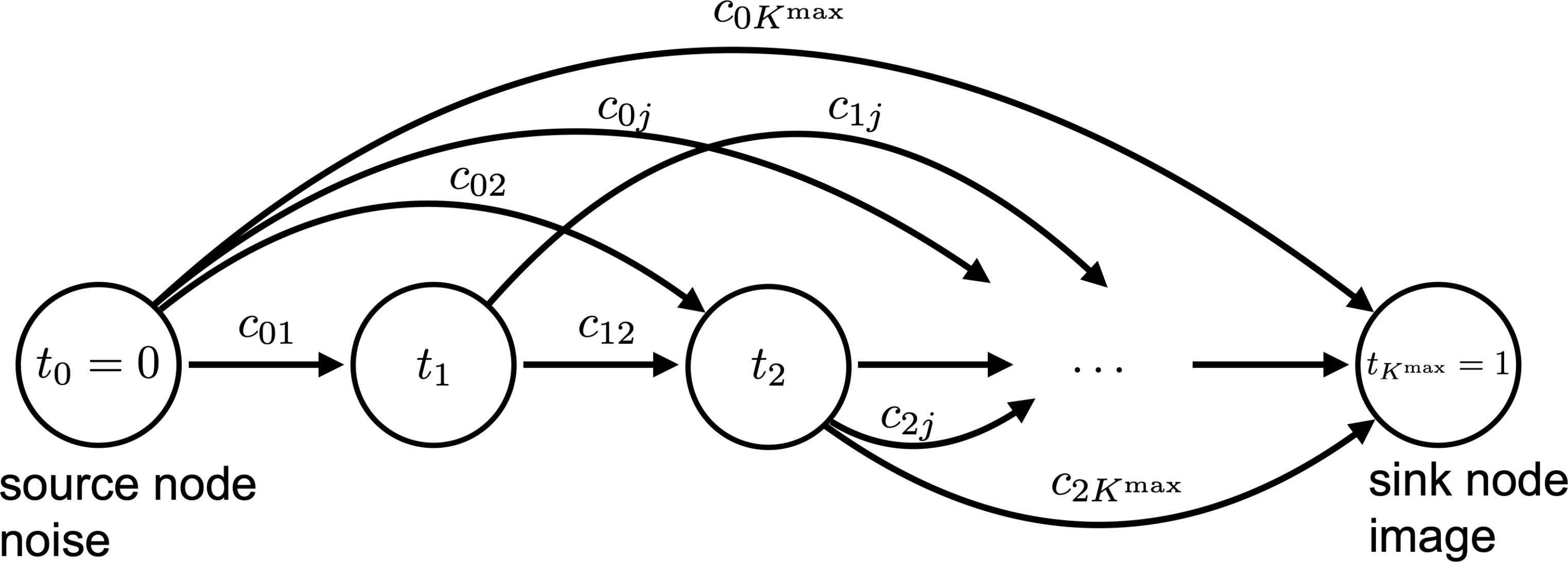}
        \caption{A network flow formulation to find the optimal sampling schedule for image generation. Time $t_0=0$ represents noise, while $t_{K^{\max}} = 1$ is the terminal data (images). Each discretized timestamp is represented by a node, with edges reflecting the one-dimensional flow of time from noise to image. The cost $c_{jk}$ associated with each edge is the sampling error estimate, measured by the average difference between the Euler one-step and the Euler $(k-j)$-step sampling between $t_j$ and $t_k$, see Section~\ref{sec:cost}.}
        \label{fig:integer}
    \end{figure}
    
\subsection{Dynamic Programming Algorithm} \label{sec:bellman}

While the integer programming problem~\eqref{eq:prob} can be solved using commercial solvers such as GUROBI or using network flow algorithms (see~\citet[\S6]{ref:skiena08algo} for an example), there are practical cases in which we need to find optimal paths for \textit{multiple} values of the budget $K$ NFEs. A convenient way to address this computation is to leverage a dynamic programming formulation, which successively builds up the error-to-go function at each node and for each number of remaining NFEs. To this end, for any timestamp $t_{\widehat j}$ and any number of remaining NFEs $\widehat k \in \llbracket 1, K^{\max} \rrbracket$, we define the error-to-go function as
\begin{equation}
        V({\color{blue}{\widehat j}}, {\color{red}{\widehat k}}) \Let \left\{
            \begin{array}{cll}
                \min & \displaystyle \sum_{j={\color{blue}{\widehat j}} }^{K^{\text{max}}-1} \sum_{k=j+1}^{K^{\text{max}}} c_{jk} z_{jk}  \\ [1.5ex]
                \st &  \sum_{j= {\color{blue}{\widehat j}}}^{K^{\text{max}}-1} \sum_{k=j+1}^{K^{\text{max}}} z_{jk} = \color{red}{\widehat k} \\ [1.5ex]
        & \sum_{k={\color{blue}{\widehat j}}+1}^{K^{\text{max}}} z_{{\color{blue}{\widehat j}}k} = 1, \quad \sum_{j={\color{blue}{\widehat j}}}^{K^{\text{max}}-1} z_{jK^{\text{max}}} = 1 \\ [1.5ex]
     &\sum_{k={\color{blue}{\widehat j}}}^{j-1} z_{kj} = \sum_{k=j+1}^{K^{\text{max}}} z_{jk} &\forall j \in  \llbracket {\color{blue}{\widehat j}}+1, K^{\text{max}-1}\rrbracket \\
     & z_{jk} \in \{0, 1\} & \forall {\color{blue}{\widehat j}} \le j < k \le  K^{\max}.
            \end{array}
        \right.
    \end{equation}
    The error-to-go $V({\color{blue}{\widehat j}}, {\color{red}{\widehat k}})$ is the minimal sampling error accumulated from time $t_{{\color{blue}{\widehat j}}}$ to the terminal time $t_{K^{\max}} = 1$ using exactly $\color{red}{\widehat k}$ NFEs. It is easy to see that the optimal value of problem~\eqref{eq:prob} is equal to $V(0, K)$. At initialization, we set the base case
    \[
    V(\widehat j, 1) = c_{\widehat j K^{\max}} \quad \forall~ \widehat j \in \llbracket 1, K^{\max} \rrbracket.
    \]
    The dynamic programming update step rolls backward following 
    \be \label{eq:dynamic}
    \forall k = 2, \ldots, K^{\max}: \qquad V(j, k) = \min_{j < \widehat j \le K^{\max}} c_{j \widehat j} + V(\widehat j, k-1).
    \ee
    The output of the dynamic programming algorithm is the error function $V$, and one can assess the Bellman optimal schedule by tracing the minimizing path following~\eqref{eq:dynamic} for each value $K$ of NFEs.

\section{Straightening Flows with Bellman Stepsize} \label{sec:straighten}

\begin{wrapfigure}{R}{0.5\textwidth}
    \begin{minipage}{1\linewidth}
    \vspace{-10mm}
    \begin{figure}[H]
        \includegraphics[width=1\linewidth]{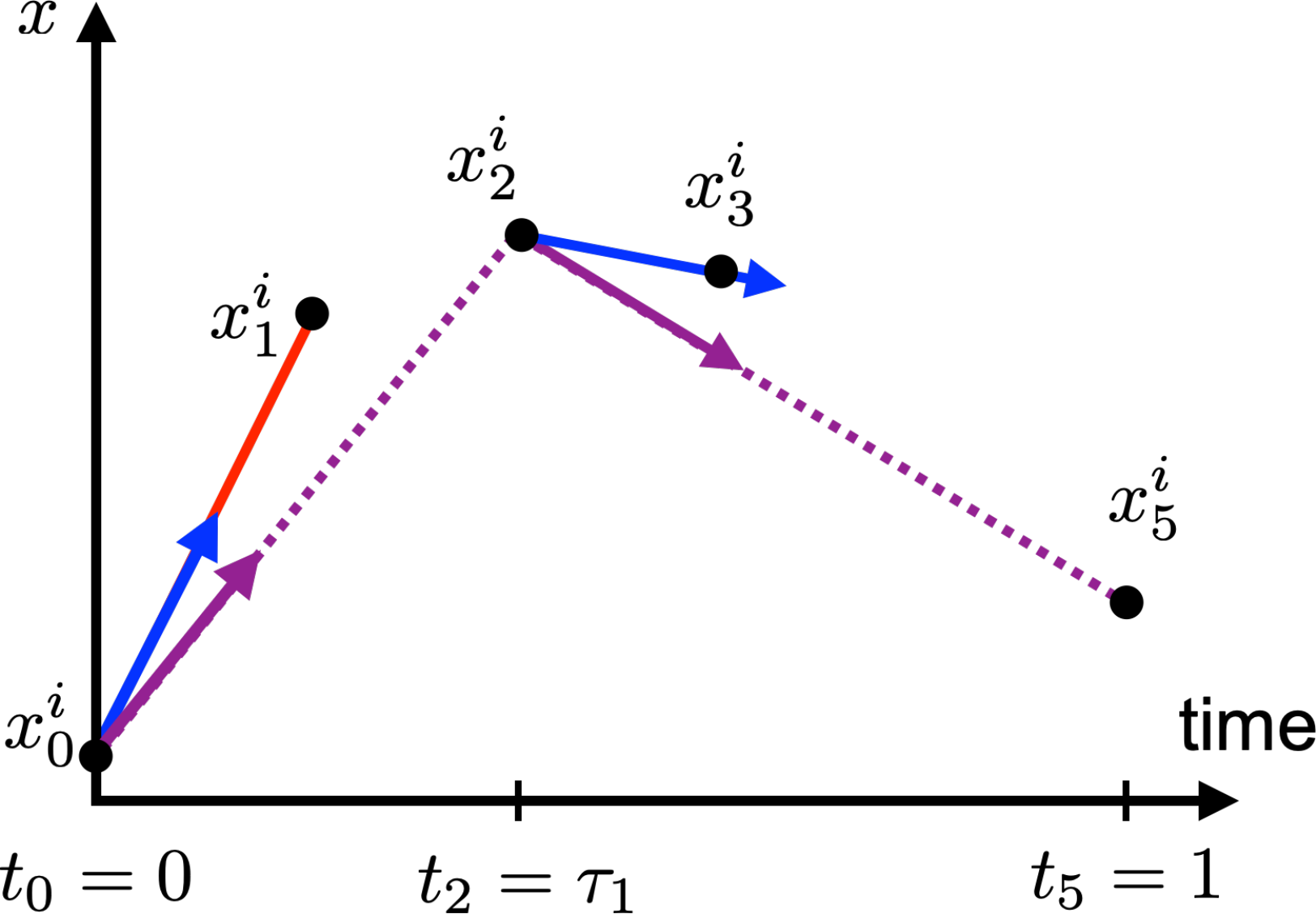}
        \caption{Continued example following Figure~\ref{fig:cost} for straightening with $K = 2$ NFEs, evaluated at time $t_0 = 0$ and time $t_2 = 0.4$. Blue arrows are velocity vectors given by the \textit{pretrained} model, and purple arrows following the dashed lines are the ideal straight path. The straightening procedure in Section~\ref{sec:straighten} aims to align the blue arrows towards the purple arrows. Arrows illustrate directions and are not drawn with proper scale.}
        \label{fig:redress}
    \end{figure}
    \vspace{-11mm}
    \end{minipage}
\end{wrapfigure}
Given the Bellman optimal stepsizes, we describe a piecewise linear straightening of the sampling curve. The straightening procedure aims to re-align the velocity network $v_\theta$ to reduce the accumulated sampling error at the terminal timestamps. For a fixed number of NFEs $K$, let $\{ \tau_0, \ldots, \tau_K\}$ be the optimal timestamps found in Section~\ref{sec:step} with $\tau_0 = 0$ and $\tau_K = 1$, which corresponds to $K$ stepsizes defined by $\tau_{k} - \tau_{k-1}$ for $k = 1, \ldots, K$.  We now modify the network weights to straighten the sampling path on each interval $[\tau_k, \tau_{k+1}]$. An intuitive explanation for the straightening procedure is illustrated in Figure~\ref{fig:redress}: here, suppose that $K = 2$, and the optimal schedule is $\{\tau_0 = 0, \tau_1 = 0.4, \tau_2 = 1\}$. For the sample $i$ drawn in Figure~\ref{fig:redress}, the Bellman sampling induces a two-piece linear path $x_0^i \to x_2^i \to x_5^i$ (dashed line). If the velocity vectors evaluated at $x_0^i$ and $x_2^i$ align with the dashed line, then the Bellman optimal Euler sampling with $K = 2$ incurs zero loss compared to the Euler uniform sampling with $K^{\max} = 5$. This motivates the following alignment procedure:
\begin{equation} \notag
    \min_{\theta}~\mathbb{E}_{X_0 \sim \pi_0}\Big[ \sum_{k = 0}^{K-1} \Big\| v_{\theta}(X_{\tau_k}^{\mc E(K^{\max})}, \tau_k) - \frac{X_{\tau_{k+1}}^{\mc E(K^{\max})} - X^{\mc E(K^{\max})}_{\tau_{k}}}{\tau_{k+1} - \tau_k} \Big\|_2^2\Big],
\end{equation}
where $X_{\tau_k}^{\mc E(K^{\max})}$ is obtained by the Euler uniform sampling with $K^{\max}$ NFEs of the initial condition $X_0 \sim \pi_0$. Replacing the expectation with $n$ empirical paths obtained by $\mc E(K^{\max})$, we have the sample averaging optimization problem for straightening
\begin{equation}
    \min_{\theta}~\frac{1}{n} \sum_{i=1}^n \sum_{k = 0}^{K-1} \| v_{\theta}(x^i_{\tau_k}, \tau_k) - \frac{x^i_{\tau_{k+1}} - x^i_{\tau_{k}}}{\tau_{k+1} - \tau_k} \|_2^2.
\end{equation}
We straighten the velocity model using a stochastic gradient descent algorithm to solve the above problem. In the main paper, we train all parameters $\theta$ of the pretrained model, whereas in Appendix~\ref{sec:lora}, we employ Low-Rank Adaptations to reduce the number of trainable parameters while preserving the performance of the straightening process.

\section{Numerical Experiments}

\label{sec:experiments}
\textbf{Settings.} We evaluate our methods on unconditioned image generation tasks. In particular,
we use the CIFAR-10 \citep{ref:krizhevsky2009learning} and three high-resolution
(256x256) datasets CelebA-HQ \citep{ref:karras2018progressive}, LSUN-Church, LSUN-Bedroom
\citep{ref:yu2015lsun}, and AFHQ-Cat. We take the checkpoints of pretrained velocity networks $v$ from the official
implementation\footnote{\url{https://github.com/gnobitab/RectifiedFlow/}} of
Rectified Flow \citep{ref:liu2022flow}, which is based on the U-Net
architecture of DDPM++ \citep{ref:song2020score}. If not mentioned otherwise, we evaluate the sample schemes with NFE=$\{4,6,8\}$. The quality of image samples is with Frechet inception distance (FID) score
\citep{ref:heusel2017gans}.

\textbf{Baselines.} A comparison between Bellman optimal stepsizes and the conventional first/second order methods using uniform stepsizes is presented in \Cref{bellman vs euler}. We also include an adaptive strategy, the Runge-Kutta method of order 5(4) from SciPy ~\citep{ref:virtanen2020scipy}.
In \Cref{ssec:redress-vs-reflow}, our finetuning procedure, Bellman Optimal Stepsize Straightening (BOSS), is compared with two baselines including the Uniform-Reflow, and Distill-k-Reflow introduced in~\citep{ref:liu2022flow}. 

\subsection{Improvements of first order and second order sampling scheme using Bellman Optimal Stepsize}
\label{bellman vs euler}

First, we benchmark pretrained Euler samplers with uniform and Bellman optimal stepsizes, calculated following \Cref{sec:bellman}. The quantitative results are demonstrated in \Cref{table:fid-sampling}. The FID is much lower for samples generated by Euler's method with the Bellman step, which shows that with Bellman's optimal step size, the generated images are of much higher quality in general. Specifically, for sampling on the three larger dimension datasets (256x256), Bellman steps can help drastically reduce FID compared to the uniform step size. This trend is also reflected in Figure~\ref{fig:three_images}, which shows our qualitative results. 

\begin{table}[H]
\centering
\caption{FID ($\downarrow$) of Euler's sampling method with uniform stepsizes vs.~Bellman optimal stepsizes on unconditional image generation task across different datasets.}
\small
\begin{tabular}{ccccccc}
\hline
\multirow{2}{*}{Dataset} & \multicolumn{2}{c}{4 NFE} & \multicolumn{2}{c}{6 NFE} & \multicolumn{2}{c}{8 NFE} \\ \cline{2-7} 
 & Euler & Bellman & Euler & Bellman & Euler & Bellman \\ \hline
CIFAR-10 & 51.95 & $\mathbf{47.57}$ & 25.69 & $\mathbf{23.35}$ & 16.82 & $\mathbf{15.74}$ \\
CelebA-HQ & 158.95 & $\mathbf{92.03}$ & 127.01 & $\mathbf{72.54}$ & 109.42 & $\mathbf{49.80}$ \\
LSUN-Church & 106.94 & $\mathbf{80.91}$ & 53.85 & $\mathbf{45.09}$ & 34.74 & $\mathbf{33.22}$ \\
AFHQ-Cat & 68.95 & $\mathbf{45.54}$ & 61.50 & $\mathbf{36.15}$ & 56.96 & $\mathbf{33.94}$ \\
\multicolumn{1}{c}{LSUN-Bedroom} & \multicolumn{1}{c}{$84.35$} & \multicolumn{1}{c}{$\mathbf{61.60}$} & \multicolumn{1}{c}{$39.19$} & \multicolumn{1}{c}{$\mathbf{35.35}$} & \multicolumn{1}{c}{$32.15$} & \multicolumn{1}{c}{$\mathbf{25.80}$} \\ \hline
\end{tabular}
\label{table:fid-sampling}
\end{table}

\begin{figure}[H]
    \centering
    \subfloat[CIFAR-10]{\includegraphics[width=0.33\linewidth]{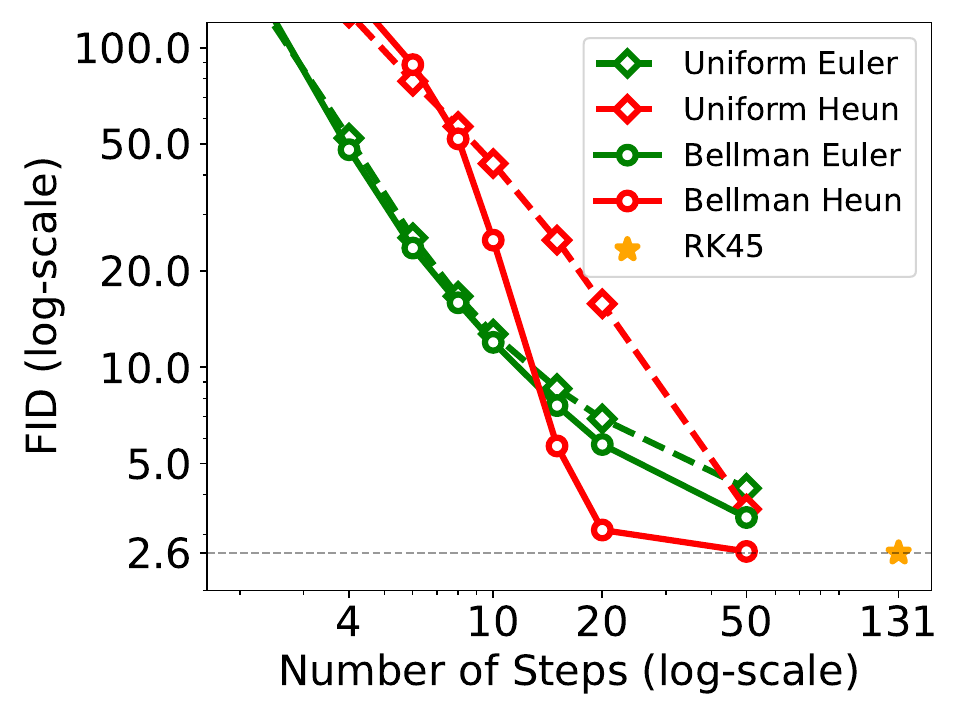}}
    \subfloat[CelebA-HQ]{\includegraphics[width=0.33\linewidth]{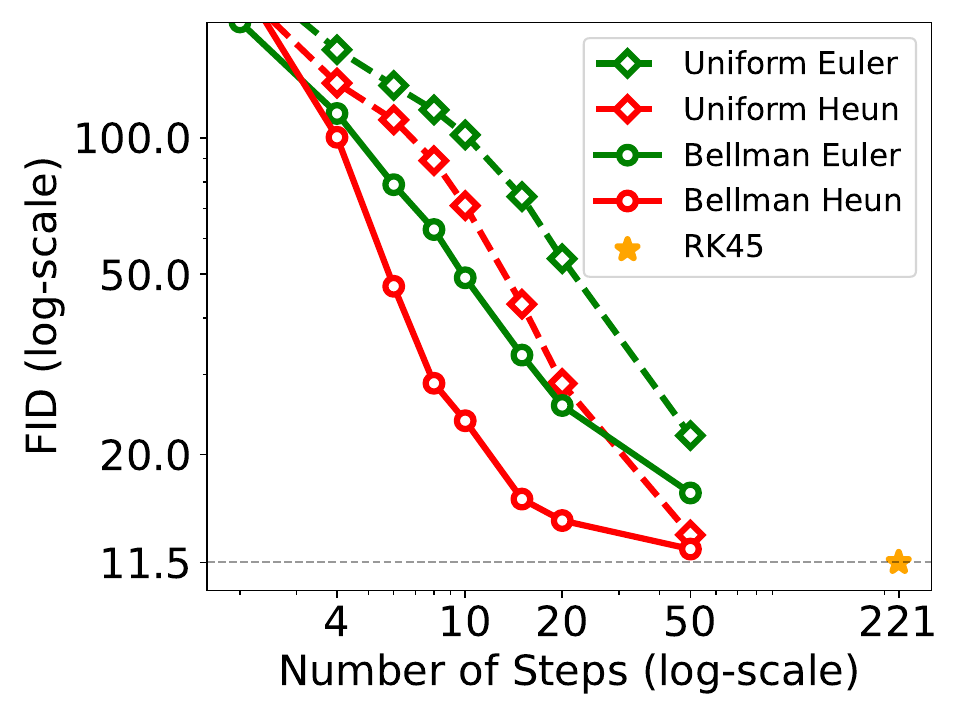}}
    \subfloat[AFHQ-Cat]{\includegraphics[width=0.33\linewidth]{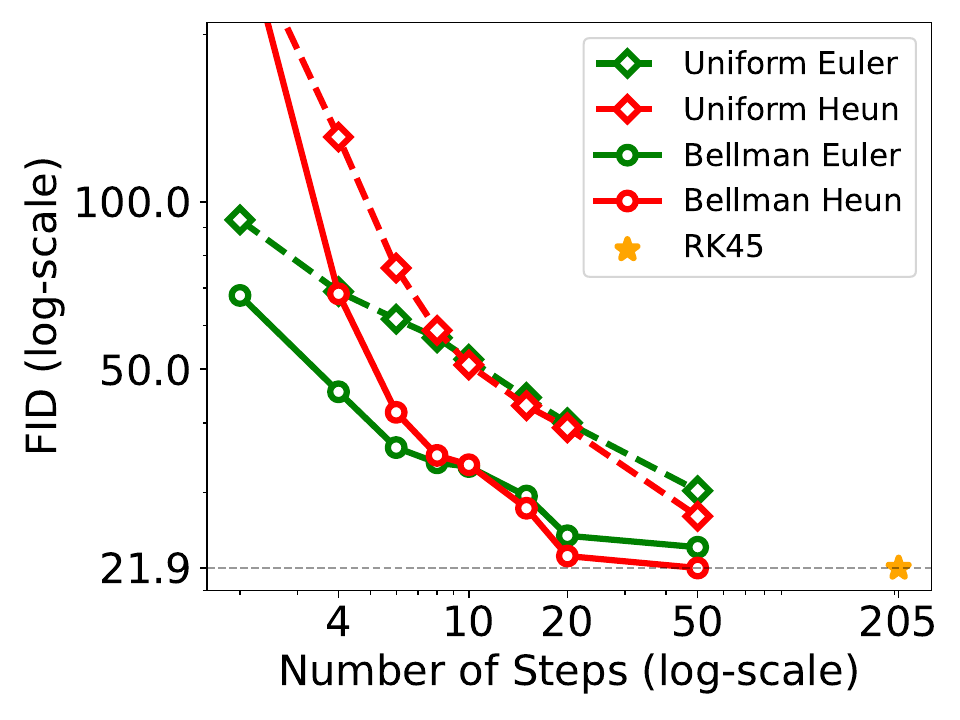}}\\
    \subfloat[LSUN-Church]{\includegraphics[width=0.33\linewidth]{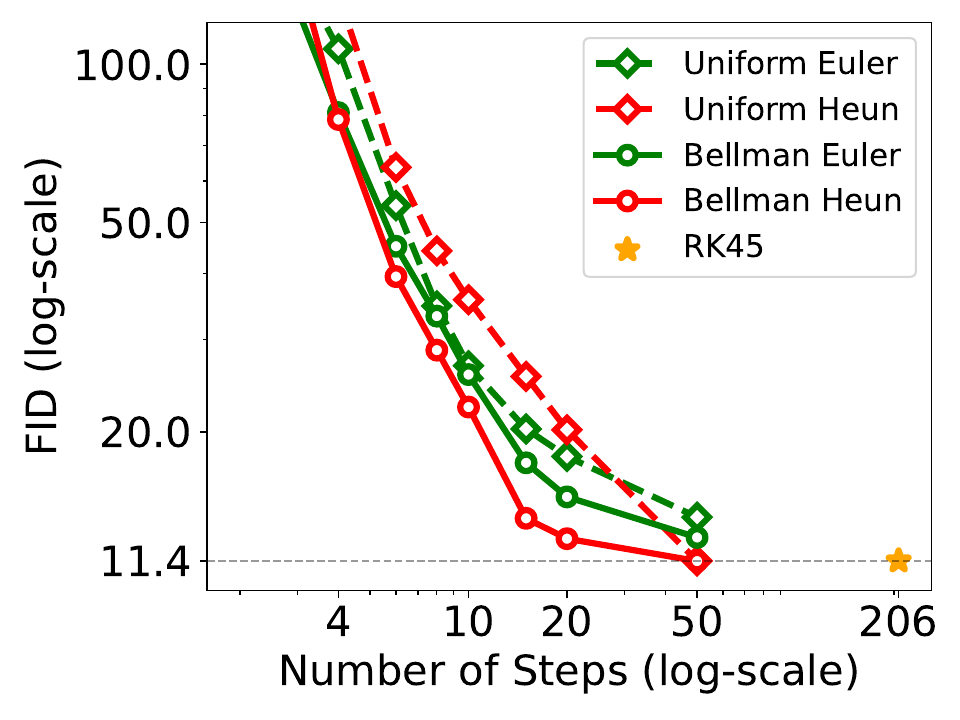}}
    \subfloat[LSUN-Bedroom]{\includegraphics[width=0.33\linewidth]{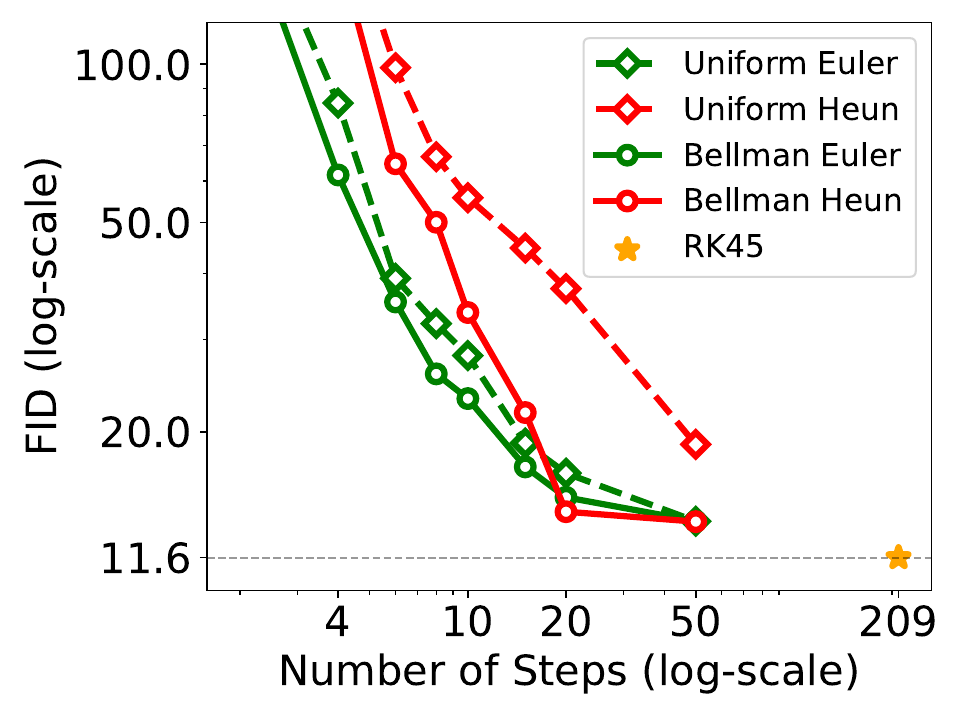}}
    \caption{The FID score of sampling methods with different numbers of function evaluations (step sizes). Images generated by samplers using Bellman stepsizes clearly show lower FID than conventional ones that use uniform step sizes. Note that Uniform Heun and Bellman Heun are second-order sampling methods that use twice the NFEs.}
    \label{fig: sampling_methods}
\end{figure}

We also empirically analyze the effect of Bellman optimal stepsizes on popular ODE solvers including Euler and Heun (second-order version). To avoid the confusion between schemes, we denote compared methods as follows:
\begin{itemize}[leftmargin=5mm]
    \item Uniform Euler and Uniform Heun are the conventional Euler and Heun methods that use uniform sampling steps.
    \item  Bellman Euler and Bellman Heun are two variants of the above methods, but using our proposed Bellman step sizes.
    \item RK45 is an adaptive strategy, the Runge-Kutta method of order 5(4) from Scipy.
\end{itemize}
The quantitative results are displayed in Figure~\ref{fig: sampling_methods}. It is evident that employing Bellman optimal stepsizes can significantly improve FID scores (image quality) compared to using uniform stepsizes across all pretrained models on four distinct datasets. The Bellman Heun methods approach the performance of RK45 with substantially fewer NFEs. To elaborate, the Bellman Heun method achieves approximately a 1\% gap compared to RK45 with just 20 sampling steps and fully recovers the performance of RK45 with 50 steps.
%\end{itemize}
%
\subsection{Effects of Reflow with Bellman Stepsize}
\label{ssec:redress-vs-reflow}
\begin{figure}[H]
\begin{tabular}{cccc}
    \includegraphics[width=0.22\columnwidth]{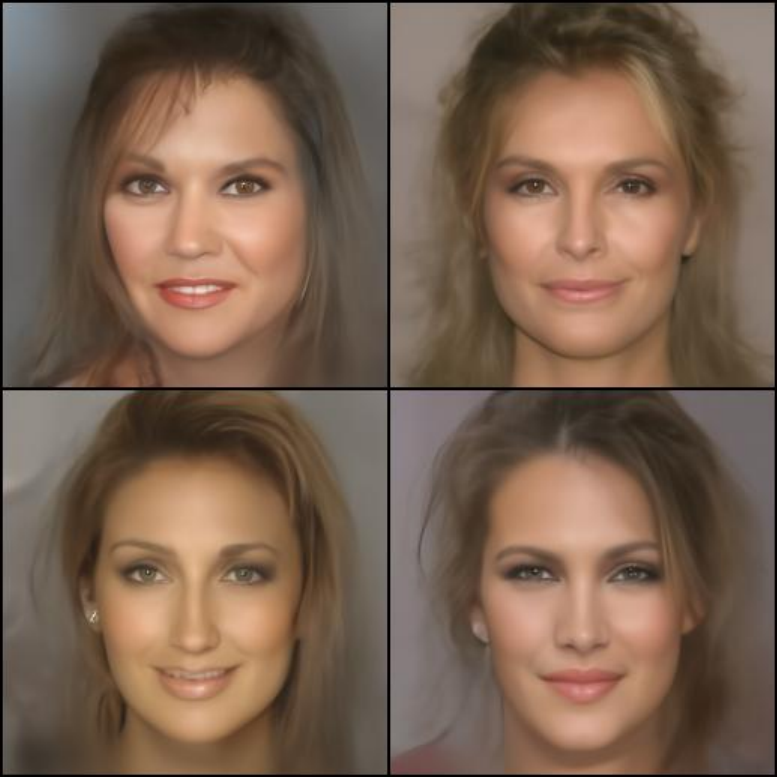}&
    \includegraphics[width=0.22\columnwidth]{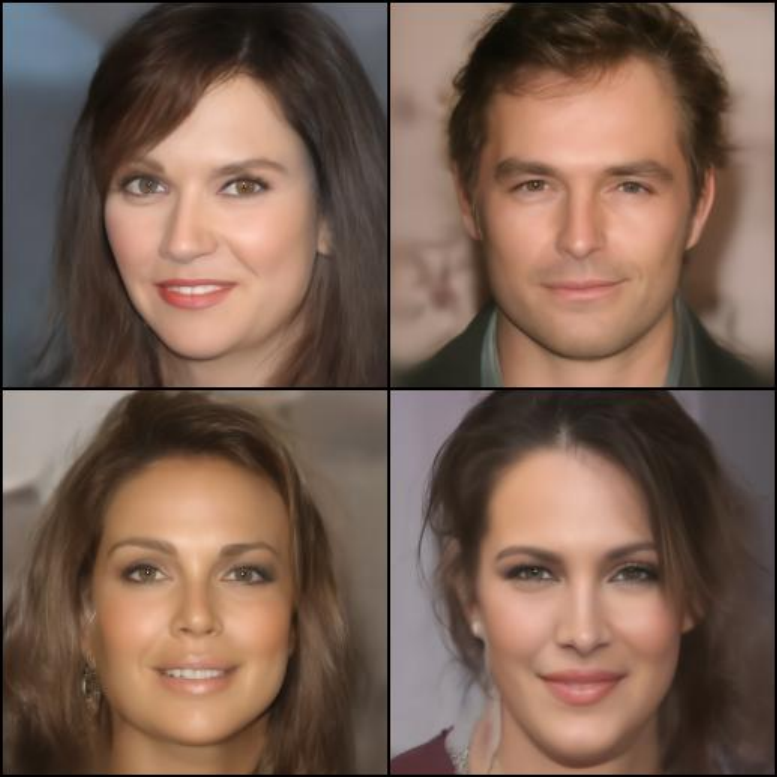}&
    \includegraphics[width=0.22\columnwidth]{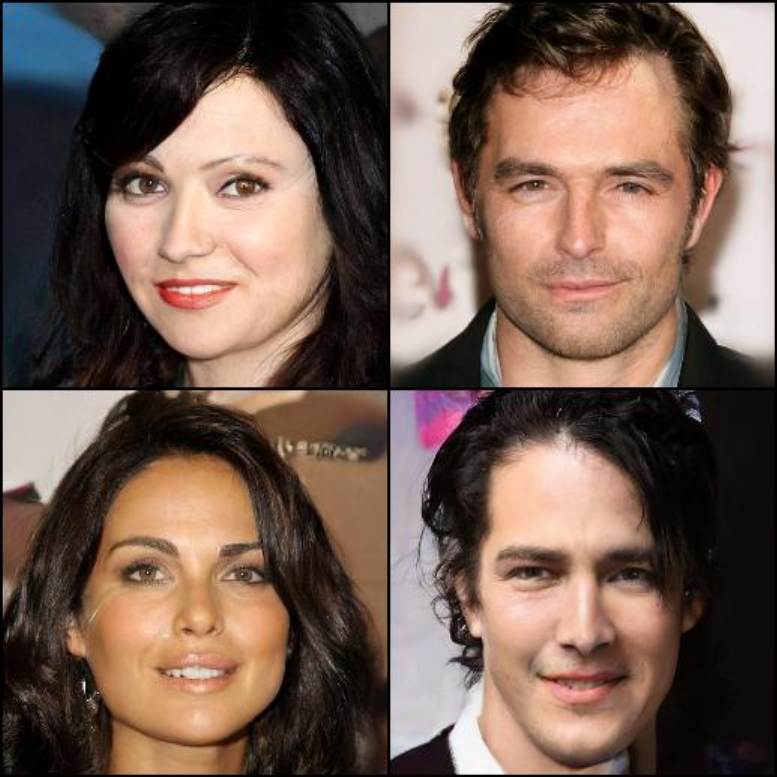}&
    \includegraphics[width=0.22\columnwidth]{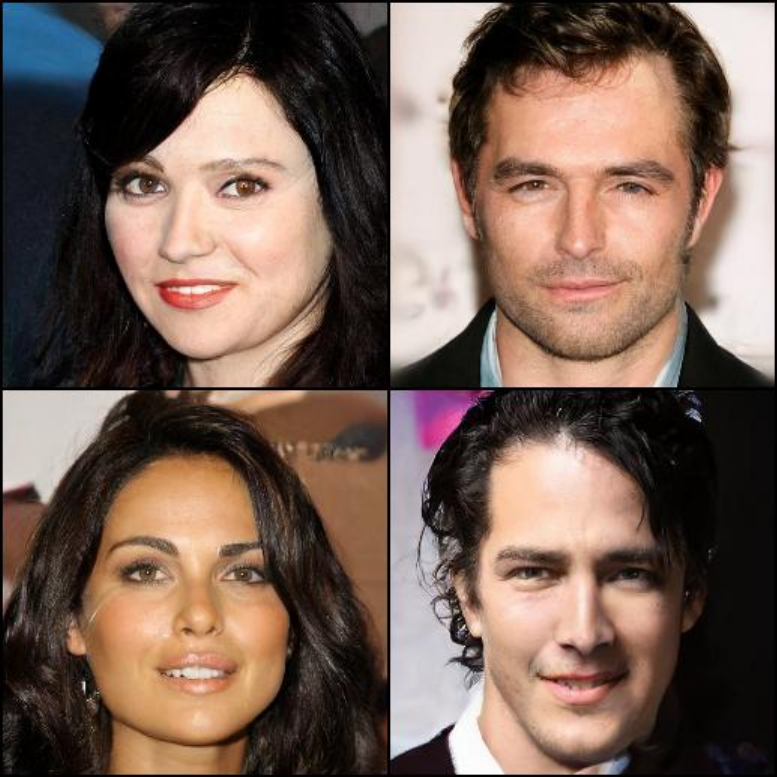}\\
    
    \includegraphics[width=0.22\columnwidth]{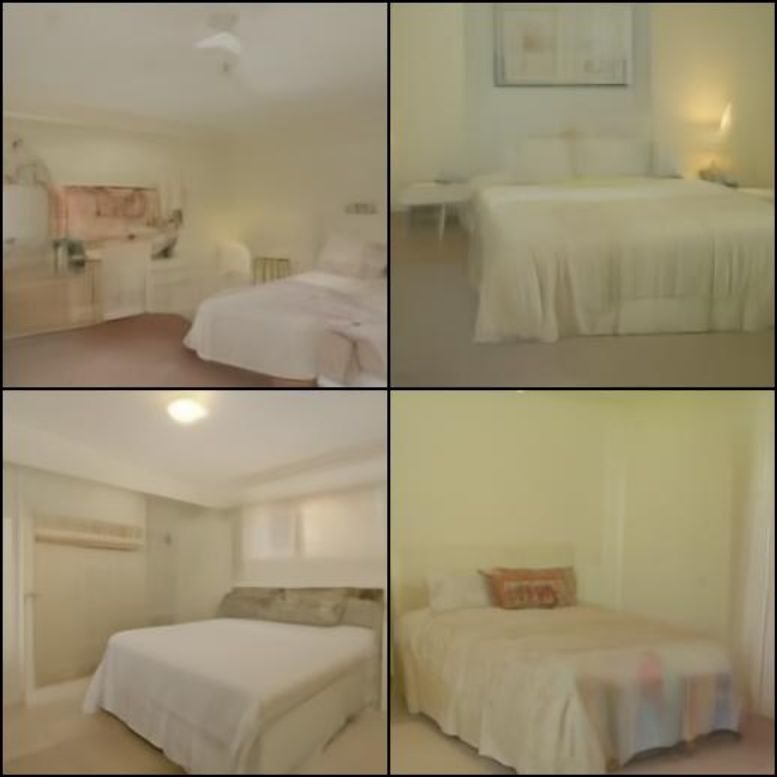}&
    \includegraphics[width=0.22\columnwidth]{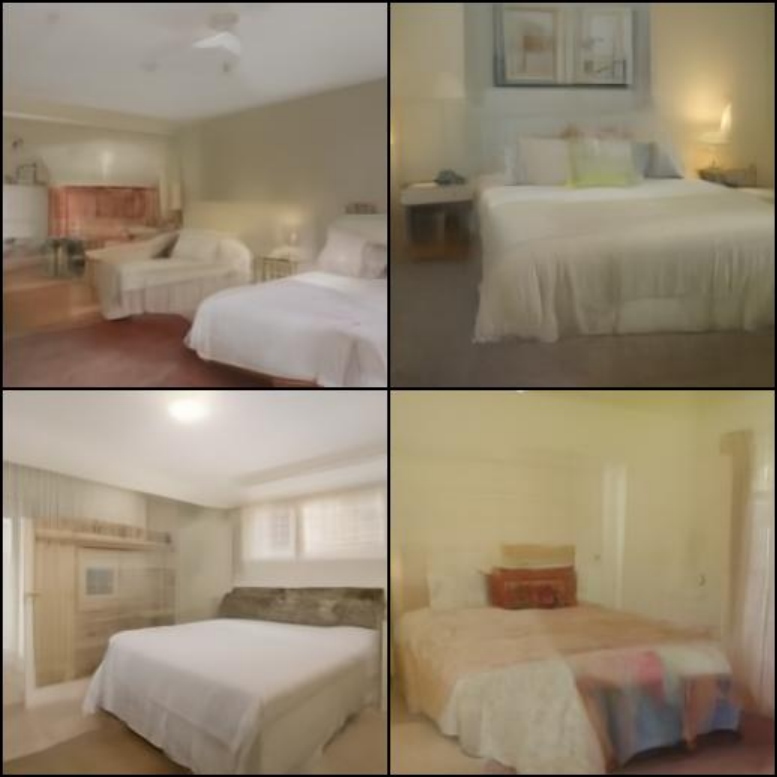}&
    \includegraphics[width=0.22\columnwidth]{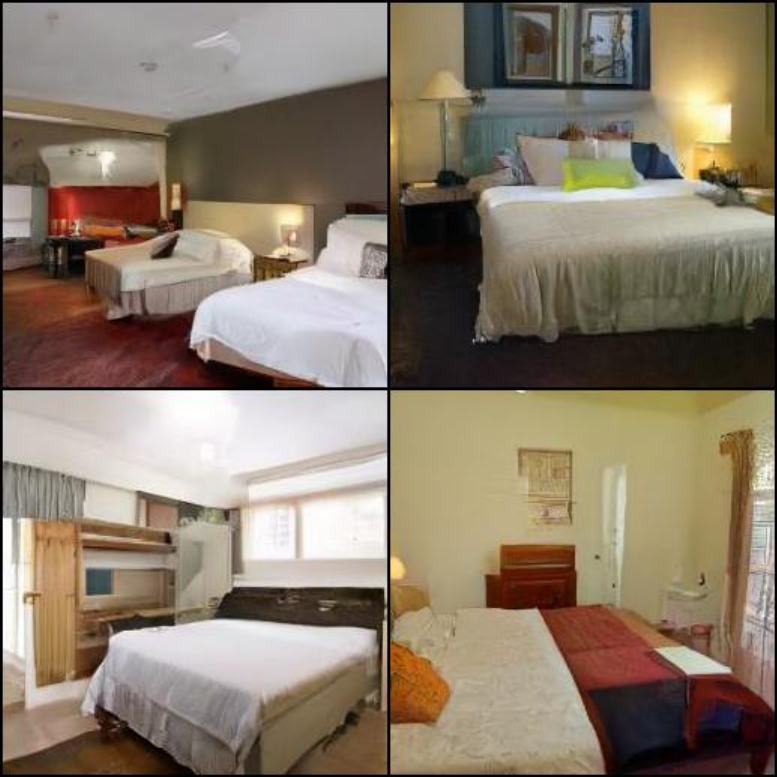}&
    \includegraphics[width=0.22\columnwidth]{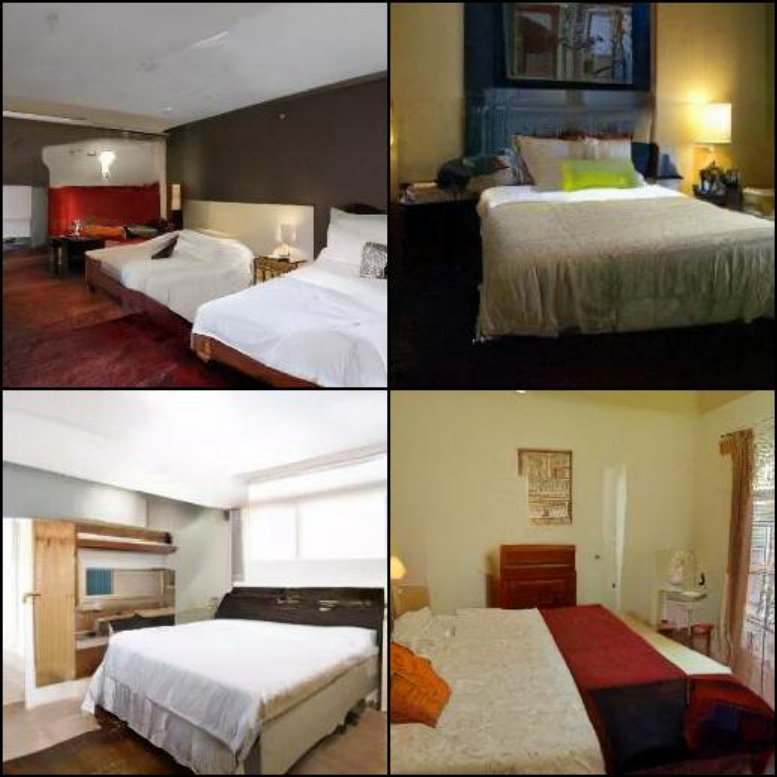}\\
    \includegraphics[width=0.22\columnwidth]{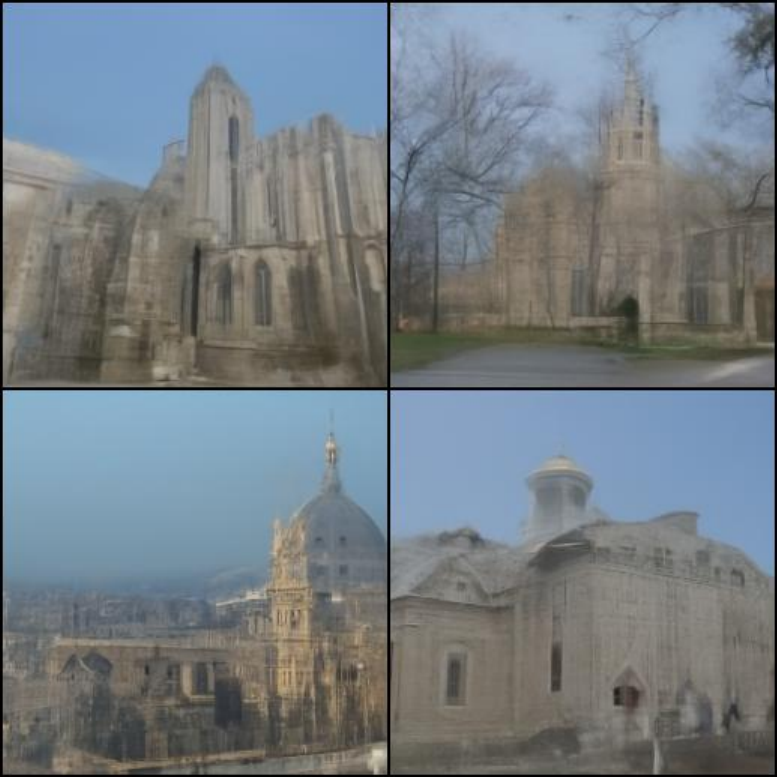}&
    \includegraphics[width=0.22\columnwidth]{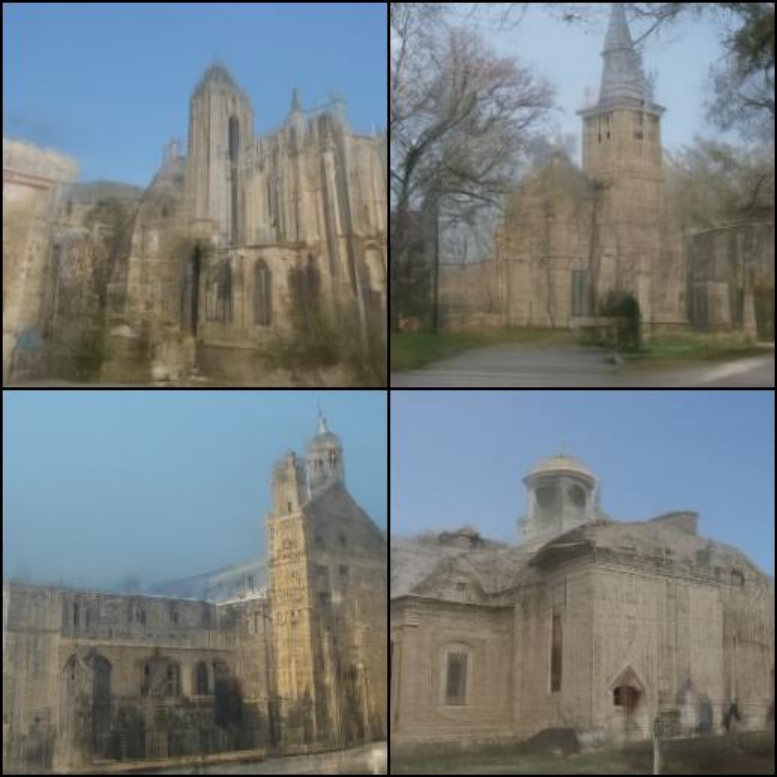}&
    \includegraphics[width=0.22\columnwidth]{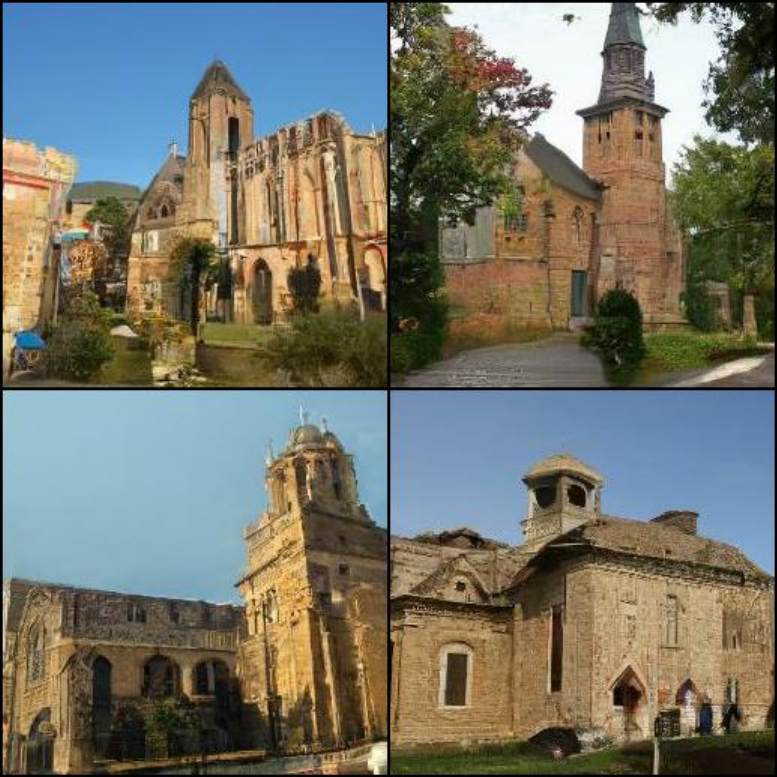}&
    \includegraphics[width=0.22\columnwidth]{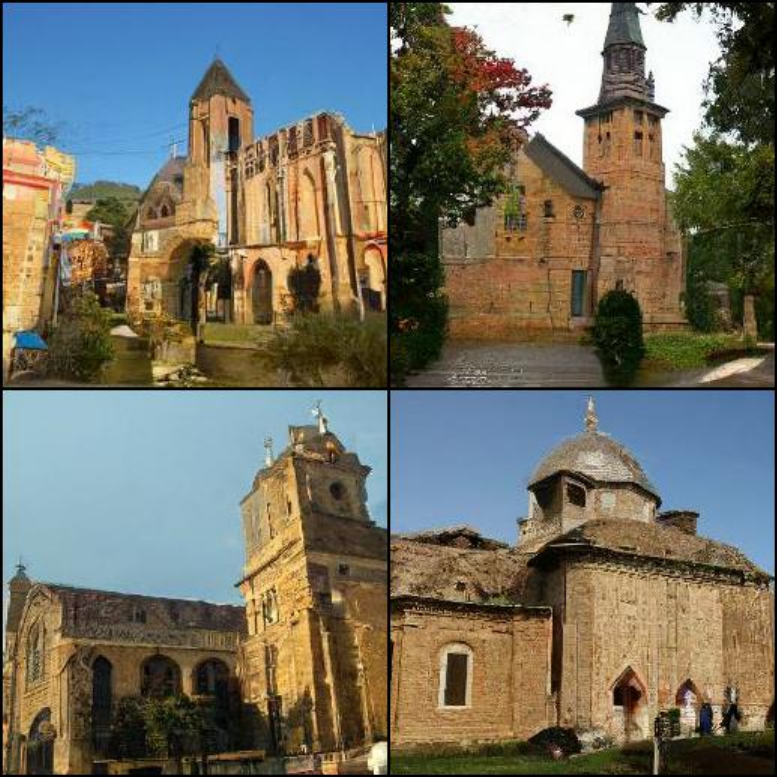}\\
    \includegraphics[width=0.22\columnwidth]{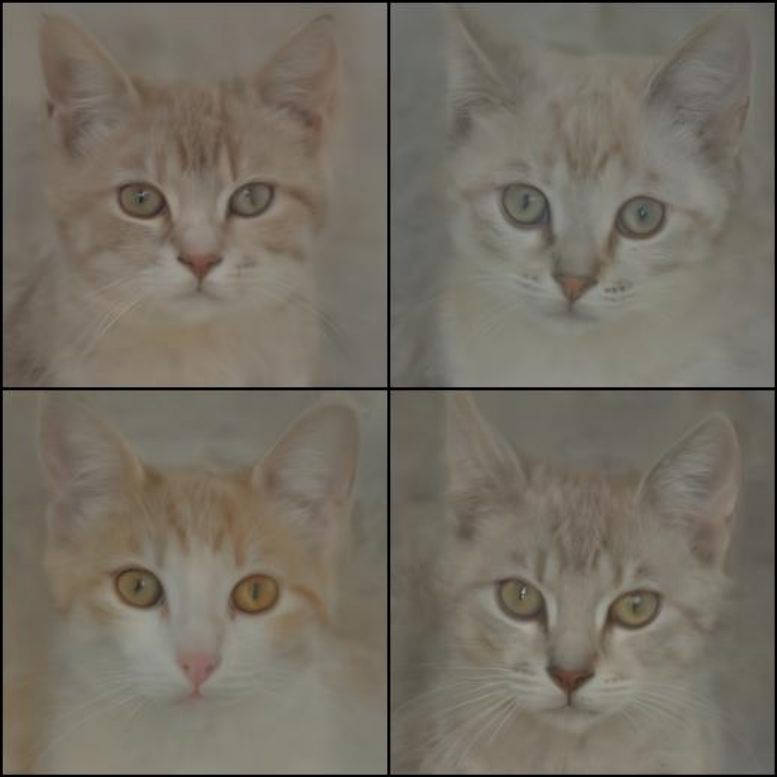}&
    \includegraphics[width=0.22\columnwidth]{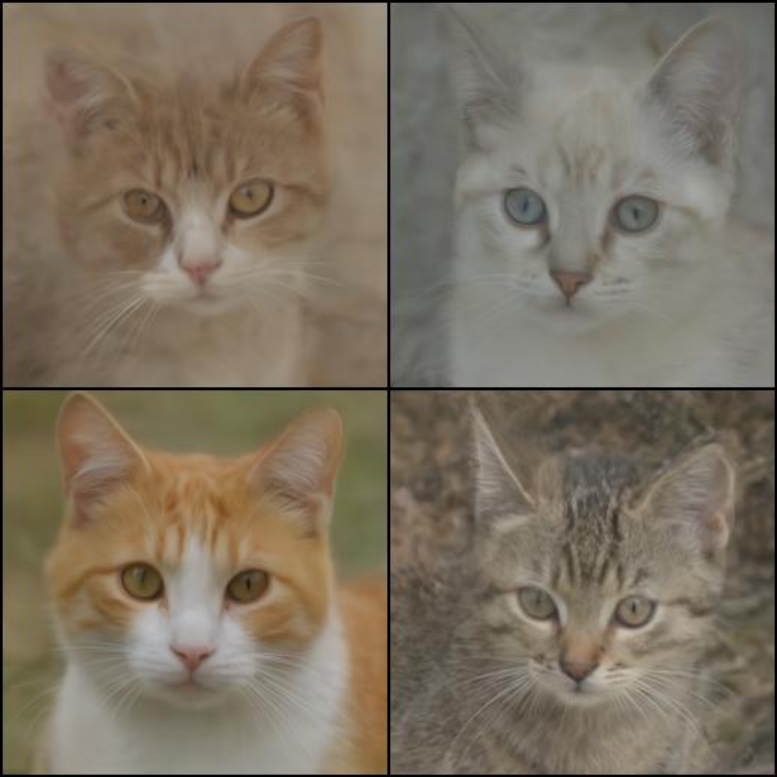}&
    \includegraphics[width=0.22\columnwidth]{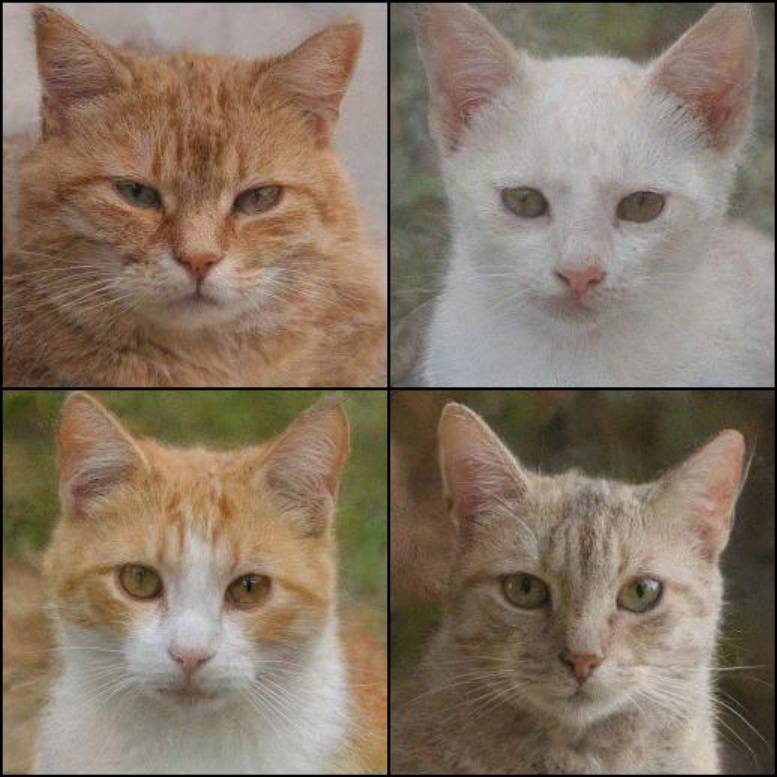}&
    \includegraphics[width=0.22\columnwidth]{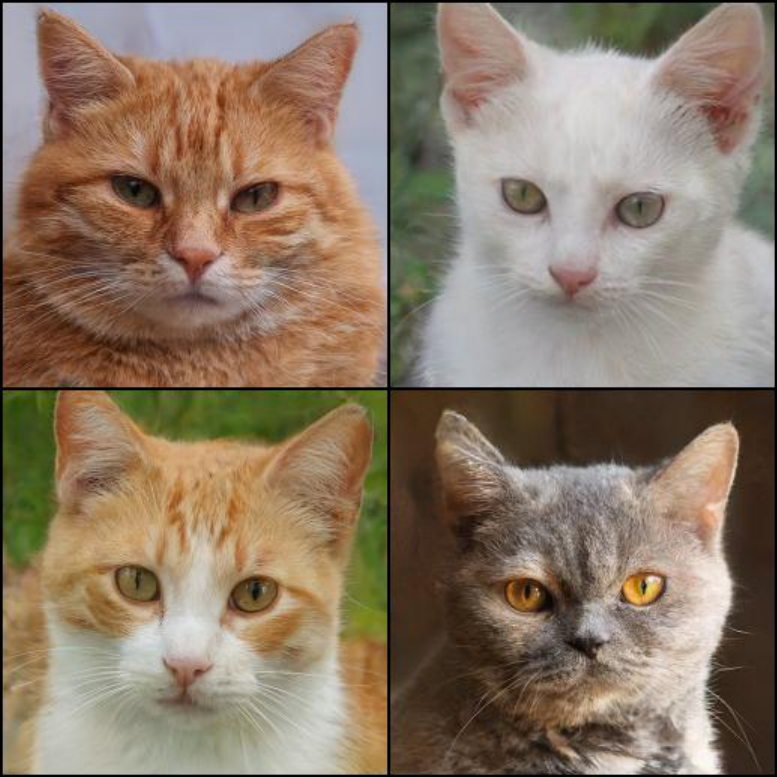}\\
(a) Euler (6 NFEs) & (b) Bellman (6 NFEs) & (c) BOSS (6 NFEs) & (d) RK45 (208 NFEs) \\
\end{tabular}
    \caption{Qualitative results on unconditional image generation task. From first to last row: CelebA-HQ/LSUN-Bedroom/LSUN-Church/AFHQ-Cat dataset. (a)-(b): Comparisons of Euler stepsizes between uniform (a) and the Bellman optimal stepsizes (b); (c)-(d): Comparisons of BOSS retraining and Runge-Kutta-45 sampling. Notice our proposed BOSS sampling has comparably similar visual quality to RK45 while requiring only 6 NFEs, compared to 208 NFEs of RK45.}
    \label{fig:three_images}
\end{figure}

After calculating the Bellman optimal step size, we follow the procedure described in \Cref{sec:straighten} to straighten the probability path. The results, seen in \Cref{table:FID-redress}, suggest that BOSS performs almost equally with the reflow procedure on CIFAR-10 but markedly better on the other four higher-dimension datasets. This is consistent with the visible improvements in sampled image quality observed in \Cref{fig:three_images}.

\begin{table}[H]
\centering
\caption{FID ($\downarrow$) of different retraining methods on the unconditional image generation task across different datasets. Distill-K-Reflow is a distillation technique that relies on the reflow of the velocity network on a discrete grid of K uniform stepsizes between 0 and 1, as elaborated in \cite{ref:liu2022flow}.}
\label{table:FID-redress}
\small
\begin{tabular}{cccc}
\hline
Dataset & Distill-6-Reflow & BOSS-6 & Uniform-Reflow \\ \hline
CIFAR-10 & $4.35$ & $4.80$ & $\mathbf{4.33}$ \\
CelebA-HQ & $34.56$ & $\mathbf{18.67}$ & $43.57$ \\
LSUN-Church & $34.52$ & $\mathbf{17.43}$ & $40.45$ \\
AFHQ-Cat & $46.24$ & $\mathbf{26.10}$ & $51.24$ \\
LSUN-Bedroom & $41.17$ & $\mathbf{18.45}$ & $45.13$ \\ \hline
\end{tabular}
\end{table}

\section{Conclusions}

This paper proposed BOSS, the Bellman Optimal stepsize Straightening method, to adapt pretrained flow-matching models under low computational resource constraints. Our method consists of two phases: first, find optimal sampling stepsizes for the pretrained model, then straighten out the velocity network on each interval of the sampling schedule. We demonstrate empirically that BOSS performs competitively in adapting pretrained models in the image generation task. Similar to training-based samplers for diffusion and flow matching models, a limitation of our method is the additional training cost to output the optimal sample step sizes. There are many potential extensions to our proposed framework to distill a guided velocity network, similar to \cite{ref:meng2023distillation}, or a computationally cheaper algorithm for calculating the Bellman sampling step sizes.

\noindent\textbf{Acknowledgments.} Viet Anh Nguyen gratefully acknowledges the generous support from the CUHK’s Improvement on Competitiveness in Hiring New Faculties Funding Scheme and the CUHK's Direct Grant Project Number 4055191. The work of Binh Nguyen is supported by the Singapore's Ministry of Education grant A-0004595-00-00. 

%\newpage
\bibliographystyle{iclr2024_conference}
\bibliography{refs}

\begin{thebibliography}{34}
\providecommand{\natexlab}[1]{#1}
\providecommand{\url}[1]{\texttt{#1}}
\expandafter\ifx\csname urlstyle\endcsname\relax
  \providecommand{\doi}[1]{doi: #1}\else
  \providecommand{\doi}{doi: \begingroup \urlstyle{rm}\Url}\fi

\bibitem[Ahuja et~al.(1993)Ahuja, Magnanti, and Orlin]{ref:ahuja1993network}
R.K. Ahuja, T.L. Magnanti, and J.B. Orlin.
\newblock \emph{Network {F}lows: {T}heory, {A}lgorithms, and {A}pplications}.
\newblock Prentice Hall, 1993.

\bibitem[Albergo \& Vanden-Eijnden(2022)Albergo and
  Vanden-Eijnden]{ref:albergo2022building}
Michael~Samuel Albergo and Eric Vanden-Eijnden.
\newblock Building normalizing flows with stochastic interpolants.
\newblock In \emph{The Eleventh International Conference on Learning
  Representations}, 2022.

\bibitem[Bao et~al.(2022)Bao, Li, Zhu, and Zhang]{ref:bao2022analytic}
Fan Bao, Chongxuan Li, Jun Zhu, and Bo~Zhang.
\newblock Analytic-{DPM}: an analytic estimate of the optimal reverse variance
  in diffusion probabilistic models.
\newblock \emph{arXiv preprint arXiv:2201.06503}, 2022.

\bibitem[Chen et~al.(2018)Chen, Rubanova, Bettencourt, and
  Duvenaud]{ref:chen2018neural}
Ricky~TQ Chen, Yulia Rubanova, Jesse Bettencourt, and David~K Duvenaud.
\newblock Neural ordinary differential equations.
\newblock \emph{Advances in Neural Information Processing Systems}, 31, 2018.

\bibitem[Heusel et~al.(2017)Heusel, Ramsauer, Unterthiner, Nessler, and
  Hochreiter]{ref:heusel2017gans}
Martin Heusel, Hubert Ramsauer, Thomas Unterthiner, Bernhard Nessler, and Sepp
  Hochreiter.
\newblock {GAN}s trained by a two time-scale update rule converge to a local
  {N}ash equilibrium.
\newblock \emph{Advances in Neural Information Processing Systems}, 30, 2017.

\bibitem[Ho et~al.(2020)Ho, Jain, and Abbeel]{ref:ho2020denoising}
Jonathan Ho, Ajay Jain, and Pieter Abbeel.
\newblock Denoising diffusion probabilistic models.
\newblock \emph{Advances in Neural Information Processing Systems},
  33:\penalty0 6840--6851, 2020.

\bibitem[Karras et~al.(2018)Karras, Aila, Laine, and
  Lehtinen]{ref:karras2018progressive}
Tero Karras, Timo Aila, Samuli Laine, and Jaakko Lehtinen.
\newblock Progressive growing of {GAN}s for improved quality, stability, and
  variation.
\newblock In \emph{International Conference on Learning Representations}, 2018.

\bibitem[Karras et~al.(2022)Karras, Aittala, Aila, and
  Laine]{ref:karras2022elucidating}
Tero Karras, Miika Aittala, Timo Aila, and Samuli Laine.
\newblock Elucidating the design space of diffusion-based generative models.
\newblock \emph{Advances in Neural Information Processing Systems},
  35:\penalty0 26565--26577, 2022.

\bibitem[Krizhevsky et~al.(2009)]{ref:krizhevsky2009learning}
Alex Krizhevsky et~al.
\newblock Learning multiple layers of features from tiny images.
\newblock \emph{cs.toronto.edu}, 2009.

\bibitem[Li et~al.(2023)Li, Li, Zheng, Wu, Xiao, Wang, Zheng, Pan, Chao, and
  Ji]{li2023autodiffusion}
Lijiang Li, Huixia Li, Xiawu Zheng, Jie Wu, Xuefeng Xiao, Rui Wang, Min Zheng,
  Xin Pan, Fei Chao, and Rongrong Ji.
\newblock Autodiffusion: Training-free optimization of time steps and
  architectures for automated diffusion model acceleration.
\newblock In \emph{Proceedings of the IEEE/CVF International Conference on
  Computer Vision}, pp.\  7105--7114, 2023.

\bibitem[Lipman et~al.(2022)Lipman, Chen, Ben-Hamu, Nickel, and
  Le]{ref:lipman2022flow}
Yaron Lipman, Ricky~TQ Chen, Heli Ben-Hamu, Maximilian Nickel, and Matthew Le.
\newblock Flow matching for generative modeling.
\newblock In \emph{The Eleventh International Conference on Learning
  Representations}, 2022.

\bibitem[Liu et~al.(2022{\natexlab{a}})Liu, Ren, Lin, and
  Zhao]{ref:liu2022pseudo}
Luping Liu, Yi~Ren, Zhijie Lin, and Zhou Zhao.
\newblock Pseudo numerical methods for diffusion models on manifolds.
\newblock \emph{arXiv preprint arXiv:2202.09778}, 2022{\natexlab{a}}.

\bibitem[Liu et~al.(2022{\natexlab{b}})Liu, Gong, and Liu]{ref:liu2022flow}
Xingchao Liu, Chengyue Gong, and Qiang Liu.
\newblock Flow straight and fast: Learning to generate and transfer data with
  rectified flow.
\newblock \emph{arXiv preprint arXiv:2209.03003}, 2022{\natexlab{b}}.

\bibitem[Liu et~al.(2023)Liu, Zhang, Ma, Peng, and Liu]{ref:liu2023instaflow}
Xingchao Liu, Xiwen Zhang, Jianzhu Ma, Jian Peng, and Qiang Liu.
\newblock Instaflow: One step is enough for high-quality diffusion-based
  text-to-image generation.
\newblock \emph{arXiv preprint arXiv:2309.06380}, 2023.

\bibitem[Lu et~al.(2022)Lu, Zhou, Bao, Chen, Li, and Zhu]{ref:lu2022dpm}
Cheng Lu, Yuhao Zhou, Fan Bao, Jianfei Chen, Chongxuan Li, and Jun Zhu.
\newblock {DPM}-solver: A fast {ODE} solver for diffusion probabilistic model
  sampling in around 10 steps.
\newblock \emph{Advances in Neural Information Processing Systems},
  35:\penalty0 5775--5787, 2022.

\bibitem[Meng et~al.(2023)Meng, Rombach, Gao, Kingma, Ermon, Ho, and
  Salimans]{ref:meng2023distillation}
Chenlin Meng, Robin Rombach, Ruiqi Gao, Diederik Kingma, Stefano Ermon,
  Jonathan Ho, and Tim Salimans.
\newblock On distillation of guided diffusion models.
\newblock In \emph{Proceedings of the IEEE/CVF Conference on Computer Vision
  and Pattern Recognition}, pp.\  14297--14306, 2023.

\bibitem[Neklyudov et~al.(2023)Neklyudov, Brekelmans, Severo, and
  Makhzani]{ref:neklyudov2023action}
Kirill Neklyudov, Rob Brekelmans, Daniel Severo, and Alireza Makhzani.
\newblock Action matching: Learning stochastic dynamics from samples.
\newblock In \emph{Proceedings of the 40th International Conference on Machine
  Learning}, 2023.

\bibitem[Parmar et~al.(2022)Parmar, Zhang, and Zhu]{parmar2021cleanfid}
Gaurav Parmar, Richard Zhang, and Jun-Yan Zhu.
\newblock On aliased resizing and surprising subtleties in {GAN} evaluation.
\newblock In \emph{Proceedings of the IEEE/CVF Conference on Computer Vision
  and Pattern Recognition}, 2022.

\bibitem[Ramesh et~al.(2022)Ramesh, Dhariwal, Nichol, Chu, and
  Chen]{ref:ramesh2022hierarchical}
Aditya Ramesh, Prafulla Dhariwal, Alex Nichol, Casey Chu, and Mark Chen.
\newblock Hierarchical text-conditional image generation with {CLIP} latents.
\newblock \emph{arXiv preprint arxiv:2204.06125}, 7, 2022.

\bibitem[Rombach et~al.(2022)Rombach, Blattmann, Lorenz, Esser, and
  Ommer]{ref:rombach2022high}
Robin Rombach, Andreas Blattmann, Dominik Lorenz, Patrick Esser, and Bj{\"o}rn
  Ommer.
\newblock High-resolution image synthesis with latent diffusion models.
\newblock In \emph{Proceedings of the IEEE/CVF Conference on Computer Vision
  and Pattern Recognition}, pp.\  10684--10695, 2022.

\bibitem[Saharia et~al.(2022)Saharia, Chan, Saxena, Li, Whang, Denton,
  Ghasemipour, Gontijo~Lopes, Karagol~Ayan, Salimans,
  et~al.]{ref:saharia2022photorealistic}
Chitwan Saharia, William Chan, Saurabh Saxena, Lala Li, Jay Whang, Emily~L
  Denton, Kamyar Ghasemipour, Raphael Gontijo~Lopes, Burcu Karagol~Ayan, Tim
  Salimans, et~al.
\newblock Photorealistic text-to-image diffusion models with deep language
  understanding.
\newblock \emph{Advances in Neural Information Processing Systems},
  35:\penalty0 36479--36494, 2022.

\bibitem[Salimans \& Ho(2021)Salimans and Ho]{ref:salimans2021progressive}
Tim Salimans and Jonathan Ho.
\newblock Progressive distillation for fast sampling of diffusion models.
\newblock In \emph{International Conference on Learning Representations}, 2021.

\bibitem[Skiena(2008)]{ref:skiena08algo}
Steven~S. Skiena.
\newblock \emph{The Algorithm Design Manual}.
\newblock Springer, 2008.

\bibitem[Sohl-Dickstein et~al.(2015)Sohl-Dickstein, Weiss, Maheswaranathan, and
  Ganguli]{ref:sohl2015deep}
Jascha Sohl-Dickstein, Eric Weiss, Niru Maheswaranathan, and Surya Ganguli.
\newblock Deep unsupervised learning using nonequilibrium thermodynamics.
\newblock In \emph{International conference on machine learning}, pp.\
  2256--2265. PMLR, 2015.

\bibitem[Song et~al.(2020{\natexlab{a}})Song, Meng, and
  Ermon]{ref:song2020denoising}
Jiaming Song, Chenlin Meng, and Stefano Ermon.
\newblock Denoising diffusion implicit models.
\newblock In \emph{International Conference on Learning Representations},
  2020{\natexlab{a}}.

\bibitem[Song et~al.(2020{\natexlab{b}})Song, Sohl-Dickstein, Kingma, Kumar,
  Ermon, and Poole]{ref:song2020score}
Yang Song, Jascha Sohl-Dickstein, Diederik~P Kingma, Abhishek Kumar, Stefano
  Ermon, and Ben Poole.
\newblock Score-based generative modeling through stochastic differential
  equations.
\newblock In \emph{International Conference on Learning Representations},
  2020{\natexlab{b}}.

\bibitem[Song et~al.(2023)Song, Dhariwal, Chen, and
  Sutskever]{ref:song23consistency}
Yang Song, Prafulla Dhariwal, Mark Chen, and Ilya Sutskever.
\newblock Consistency models.
\newblock In \emph{Proceedings of the 40th International Conference on Machine
  Learning}, volume 202 of \emph{Proceedings of Machine Learning Research},
  pp.\  32211--32252. PMLR, 23--29 Jul 2023.

\bibitem[Tachibana et~al.(2021)Tachibana, Go, Inahara, Katayama, and
  Watanabe]{ref:tachibana2021quasi}
Hideyuki Tachibana, Mocho Go, Muneyoshi Inahara, Yotaro Katayama, and Yotaro
  Watanabe.
\newblock Quasi-{T}aylor samplers for diffusion generative models based on
  ideal derivatives.
\newblock \emph{arXiv preprint arXiv:2112.13339}, 2021.

\bibitem[Virtanen et~al.(2020)Virtanen, Gommers, Oliphant, Haberland, Reddy,
  Cournapeau, Burovski, Peterson, Weckesser, Bright,
  et~al.]{ref:virtanen2020scipy}
Pauli Virtanen, Ralf Gommers, Travis~E Oliphant, Matt Haberland, Tyler Reddy,
  David Cournapeau, Evgeni Burovski, Pearu Peterson, Warren Weckesser, Jonathan
  Bright, et~al.
\newblock Sci{P}y 1.0: {F}undamental algorithms for scientific computing in
  {P}ython.
\newblock \emph{Nature Methods}, 17\penalty0 (3):\penalty0 261--272, 2020.

\bibitem[Wang et~al.(2023)Wang, Wang, Dinh, Du, and Xu]{wang2023learning}
Yunke Wang, Xiyu Wang, Anh-Dung Dinh, Bo~Du, and Charles Xu.
\newblock Learning to schedule in diffusion probabilistic models.
\newblock In \emph{Proceedings of the 29th ACM SIGKDD Conference on Knowledge
  Discovery and Data Mining}, pp.\  2478--2488, 2023.

\bibitem[Watson et~al.(2021)Watson, Ho, Norouzi, and Chan]{watson2021learning}
Daniel Watson, Jonathan Ho, Mohammad Norouzi, and William Chan.
\newblock Learning to efficiently sample from diffusion probabilistic models.
\newblock \emph{arXiv preprint arXiv:2106.03802}, 2021.

\bibitem[Yu et~al.(2015)Yu, Seff, Zhang, Song, Funkhouser, and
  Xiao]{ref:yu2015lsun}
Fisher Yu, Ari Seff, Yinda Zhang, Shuran Song, Thomas Funkhouser, and Jianxiong
  Xiao.
\newblock Lsun: Construction of a large-scale image dataset using deep learning
  with humans in the loop.
\newblock \emph{arXiv preprint arXiv:1506.03365}, 2015.

\bibitem[Zhang \& Chen(2022)Zhang and Chen]{ref:zhang2022fast}
Qinsheng Zhang and Yongxin Chen.
\newblock Fast sampling of diffusion models with exponential integrator.
\newblock In \emph{The Eleventh International Conference on Learning
  Representations}, 2022.

\bibitem[Zheng et~al.(2023)Zheng, Lu, Chen, and Zhu]{ref:zheng2023dpm}
Kaiwen Zheng, Cheng Lu, Jianfei Chen, and Jun Zhu.
\newblock {DPM}-solver-v3: Improved diffusion {ODE} solver with empirical model
  statistics.
\newblock In \emph{Thirty-seventh Conference on Neural Information Processing
  Systems}, 2023.

\end{thebibliography}
\newpage
\appendix

\section{Details of Experiments}

We use the following checkpoints that are downloaded from the GitHub folder \footnote{\url{https://github.com/gnobitab/RectifiedFlow/}}:
\begin{itemize}
\item CIFAR-10: at iteration 800,000;
\item CelebA-HQ: at iteration 1,000,000;
\item LSUN-Church: at iteration 1,200,000;
\item LSUN-Bedroom: at iteration 1,000,000;
\item AFHQ-Cat: at iteration 1,000,000.
\end{itemize}

The pretrained models are finetuned in 12,000 iterations. One iteration is the passing and backpropagation process for a batch including 15 samples. Due to the similar cost of training between finetuning methods, we report the average GPU hours consumed on each pretrained model up to 12000 iterations, using NVIDIA RTX A5000.
\begin{itemize}
    \item CIFAR-10: 3.56 training hours.
    \item CelebA-HQ: 10.35 training hours.
    \item LSUN-Church: 13.43 training hours.
    \item LSUN-Bedroom: 14.23 training hours.
    \item AFHQ-Cat: 9.30 training hours.
\end{itemize}
With this limited budget of resources, the proposed method, BOSS, achieves significantly better performance than other methods in terms of FID score.
The value of $N$ in Equation~\eqref{eq: local_truncation_error} and $K^{\max}$ are fixed at $100$, and $100$ in all experiments if not mentioned. This setup $N = 100$ means we only use one batch sampling to calculate the truncation errors between timestamps in equation~\eqref{eq: local_truncation_error}. It is worth noting that this setup highlights the low-cost and limited-resources requirement of our proposal.

\paragraph{FID Calculation.} All the FID metrics of related works are either cited from previous baselines or are calculated (when there are no such figures reported) based on the Clean-FID paper~\citep{parmar2021cleanfid}, which unifies the FID calculation to make a fair comparison between papers. These four datasets were downloaded following the instructions from their original papers. We then create the stats file by the clean-fid project. The FID score is calculated based on 50,000 generated images and the stats file.

\section{Description of the Dynamic Programming Algorithm}
\label{sec:dynamic programing}
This section presents the pseudocode in Algorithm~\ref{alg:min_cost_path} for the practical implementation of the dynamic programming algorithm designed to determine the Bellman optimal stepsizes. The algorithm takes a cost matrix, denoted as $c$, as input, where $c_{jk}$ is computed using Equation~\eqref{eq: local_truncation_error} and the specified number of function evaluations (NFEs). The most resource-intensive aspect of this code is the nested loop responsible for calculating $\kappa(j, k)$, incurring a time complexity of $O((K^{\max})^2 \times K)$. The choice of $K^{\max}$ is crucial, aiming for the Euler sampling method with $K^{\max}$ stepsizes to precisely replicate the trajectory of the Ordinary Differential Equation (ODE). Typically, $K^{\max}$ falls within the range of 100 to 1,000, ensuring accuracy. Given this range for $K^{\max}$ and the variable $K$ ranging from 2 to 1,000, the algorithm executes within milliseconds in all scenarios.
\begin{algorithm}[H]
	\caption{Minimum Cost Path Computation}
	\label{alg:min_cost_path}
	\begin{algorithmic}
		\STATE {\bfseries Input:} Cost matrix $c(j, k) = c_{jk} (j,k=0, \ldots, K^{\max})$ , the target NFEs $K$.
        
        \STATE Set $\kappa(i,j) \leftarrow +\infty \quad \forall 0 \leq i \leq K^{\max}, 0 \leq j \leq K$
        \STATE Set $\kappa(i, 1) \leftarrow c(i, K^{\max}) \quad \forall 0 \leq i \leq K^{\max}$
        \FOR{$k = 2 $ \TO  $K$}
            \FOR{$j = 0$ \TO $K^{\max} - 1$}
                \FOR{$i = j + 1 $ \TO $K^{\max} - 1$}
                        \STATE $\kappa(j, k) \leftarrow \min \{\kappa(j, k), c(j, i) + \kappa(i, k-1) \}$
                \ENDFOR
            \ENDFOR
        \ENDFOR
        
        \STATE Initialize $\psi \leftarrow [0]$, $\omega \leftarrow 0$
        
        \FOR{$k = K $ \TO $1$}
            \FOR{$j = \omega + 1 $ \TO $K^{\max}$}
                \IF{$\kappa(\omega, k) = c(\omega,j) + \kappa(j, k - 1)$}
                    \STATE Append $j$ to $\psi$ and set $\omega \leftarrow j$
                    \STATE \textbf{break}
                \ENDIF
            \ENDFOR
        \ENDFOR
        
        \STATE Append $K^{\max}$ to $\psi$.
        \RETURN $\psi, \kappa(0, K)$
	\end{algorithmic}
\end{algorithm}

\section{Empirical Analysis about Bellman Optimal Stepsizes}

\subsection{A Common Trend in Bellman Optimal Stepsizes for Pretrained Models on Different Datasets}
\begin{figure}[H]
    \centering
    \includegraphics[width=0.7\linewidth]{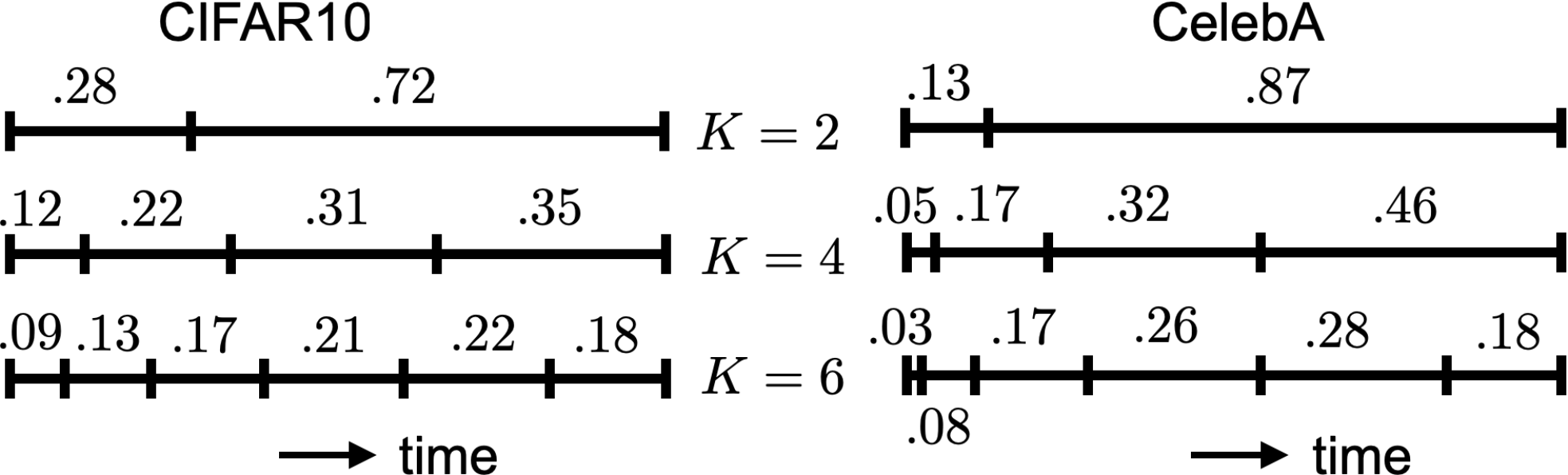}
        \caption{Optimal Bellman stepsizes for CIFAR-10 and CelebA-HQ at $K = 2, 4$ and $6$. One can identify a common pattern of smaller steps at the beginning and the end of the sampling procedure}
        \label{fig:bos}
    \end{figure}

We plot the Bellman Optimal stepsizes in \Cref{fig:bos}. It shows a common trend of sampling with small initial steps and then larger stepsizes for intermediate iterations. At $K = 6$, we can observe that the last step is smaller than the penultimate step, hinting that the sampling process aims to take smaller final steps to refine the output. This refining trend is more evident for $K=8$.

\subsection{Empirical Evidence for the Stepsizes Trend}
\label{sec:explain_steps}
This section aims to experimentally explain the trend of smaller stepsizes at the beginning and the end of the sampling procedure. Given a velocity network $v_\theta$, we empirically estimate the curvature using the following procedure:
\begin{enumerate}[leftmargin=5mm]
    \item Sample $N$ noise instances $x_0^i$ i.i.d.~from a Gaussian distribution (we set $N \approx 1000$).
    \item Forward each noise instance using $v_\theta$ to obtain $x_t^i$ for $t = 1, \ldots, K^{\max}$.
    \item Compute the local curvature for each trajectory:
    \[
    \mathrm{Curv}_t^i = \| x_t^i - \frac{x_{t+1}^i + x_{t-1}^i }{2} \|_2^2 \quad \text{for } t = 2, \ldots, K^{\max}-1.
    \]
    \item Calculate the average curvature:
    \[
        \widehat{\mathrm{Curv}}_t= \frac{1}{N} \sum_{i=1}^N \mathrm{Curv}_t^i \qquad \forall t = 2, \ldots, K^{\max} - 1.
    \]
\end{enumerate}

\begin{figure}[H]
    \centering
    \subfloat[CIFAR-10]{\includegraphics[width=0.33\linewidth]{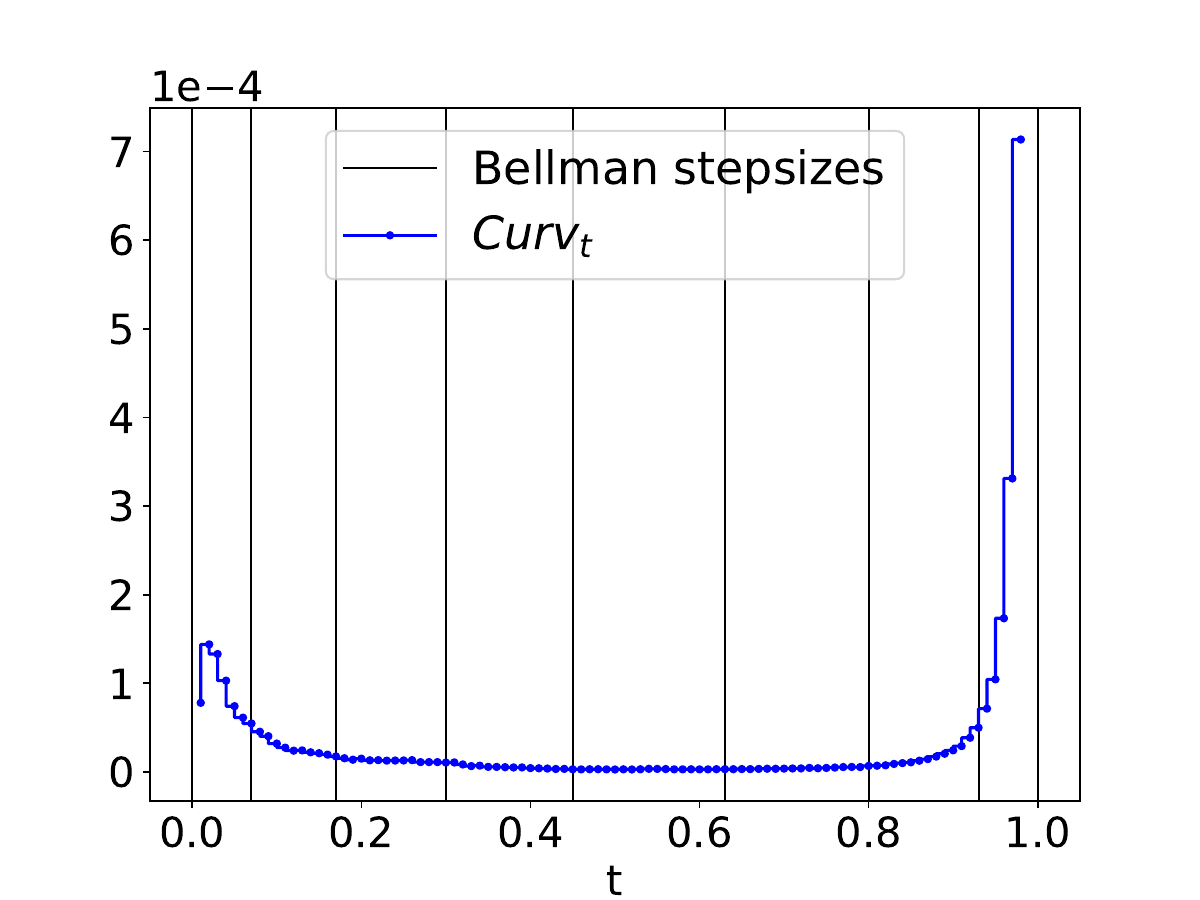}}
    \subfloat[CelebA-HQ]{\includegraphics[width=0.33\linewidth]{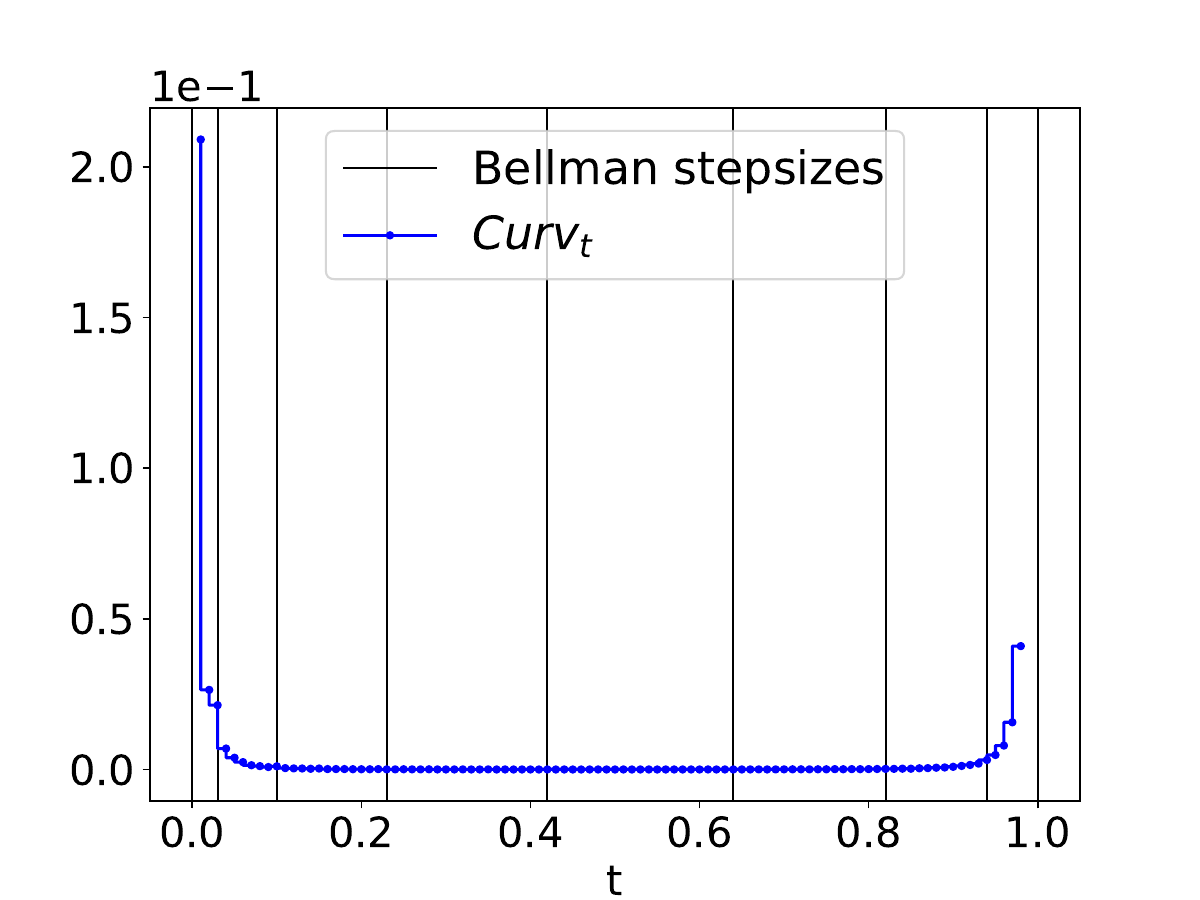}}
    \subfloat[AFHQ-Cat]{\includegraphics[width=0.33\linewidth]{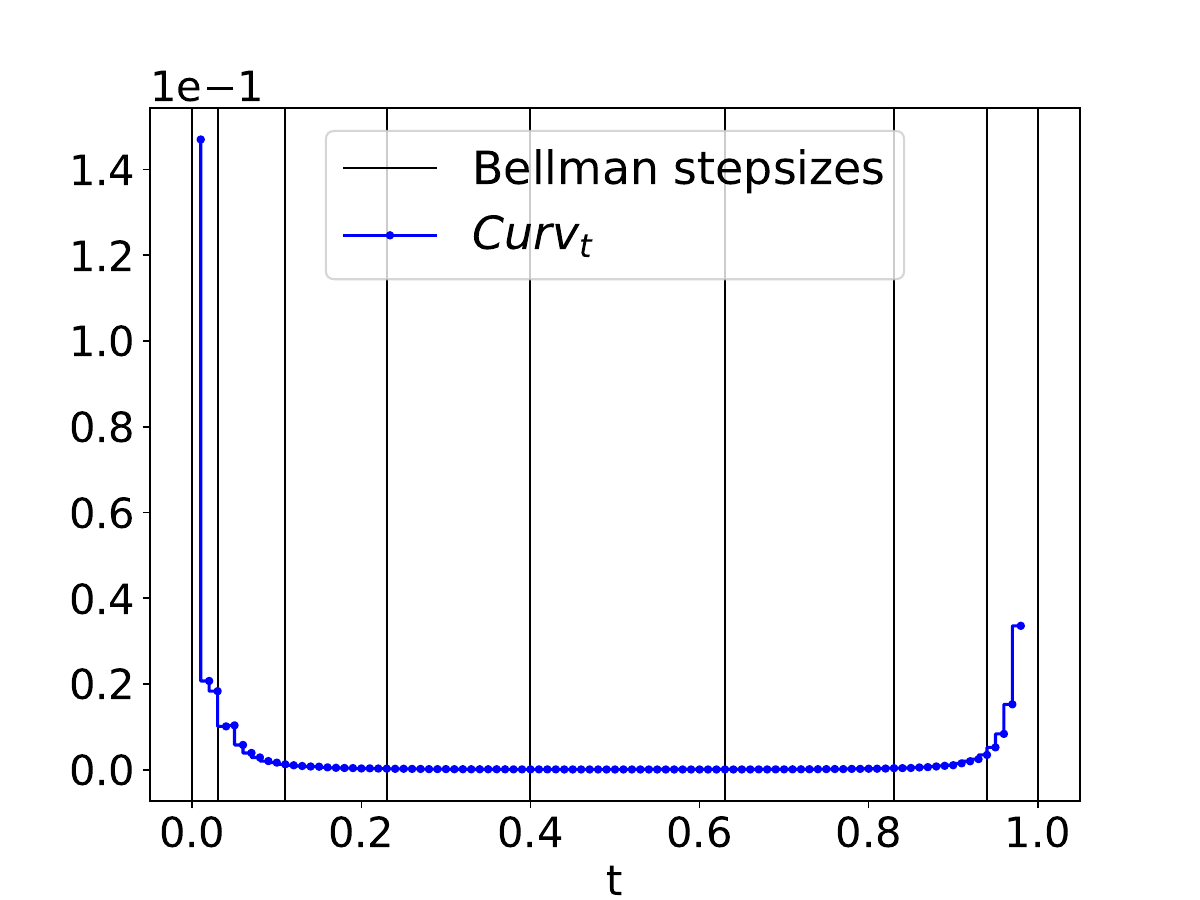}}\\
    \subfloat[LSUN-Church]{\includegraphics[width=0.33\linewidth]{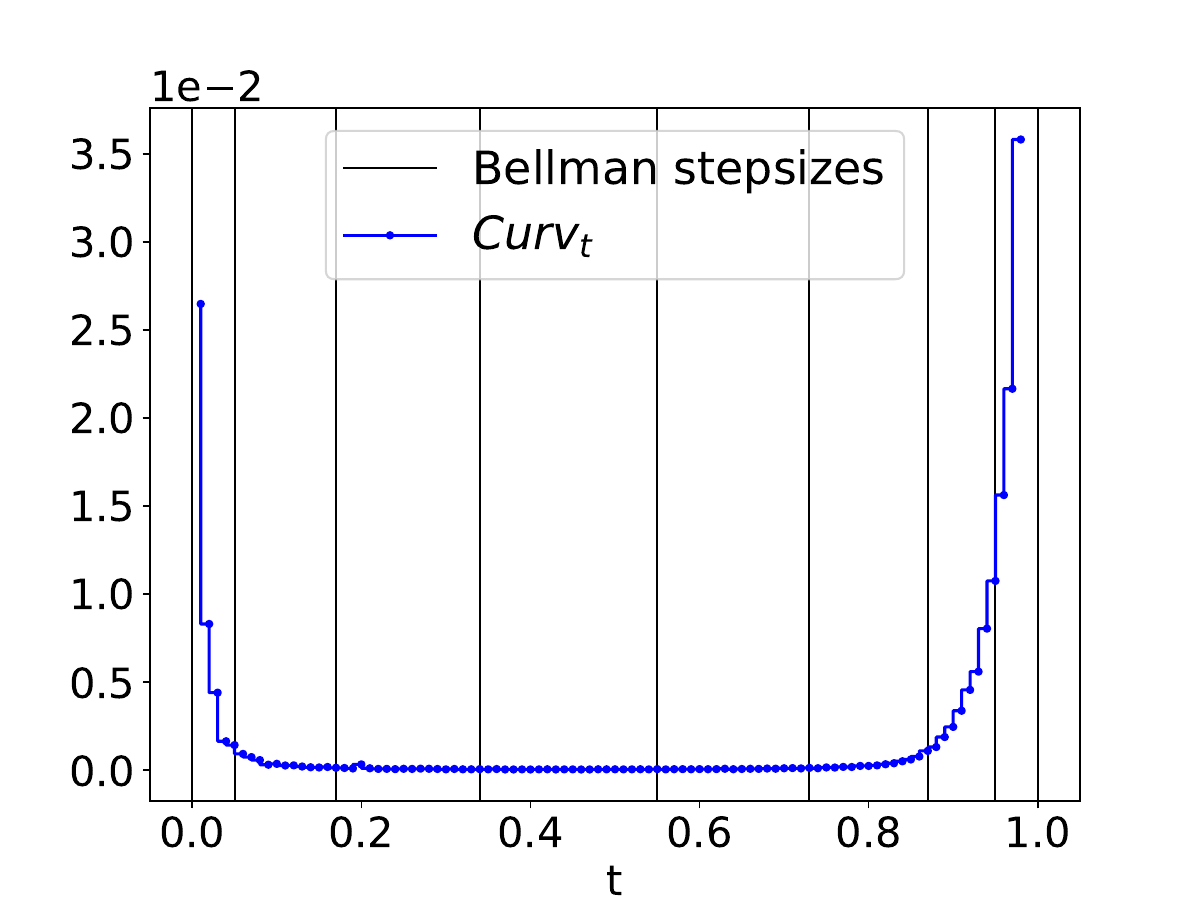}}
    \subfloat[LSUN-Bedroom]{\includegraphics[width=0.33\linewidth]{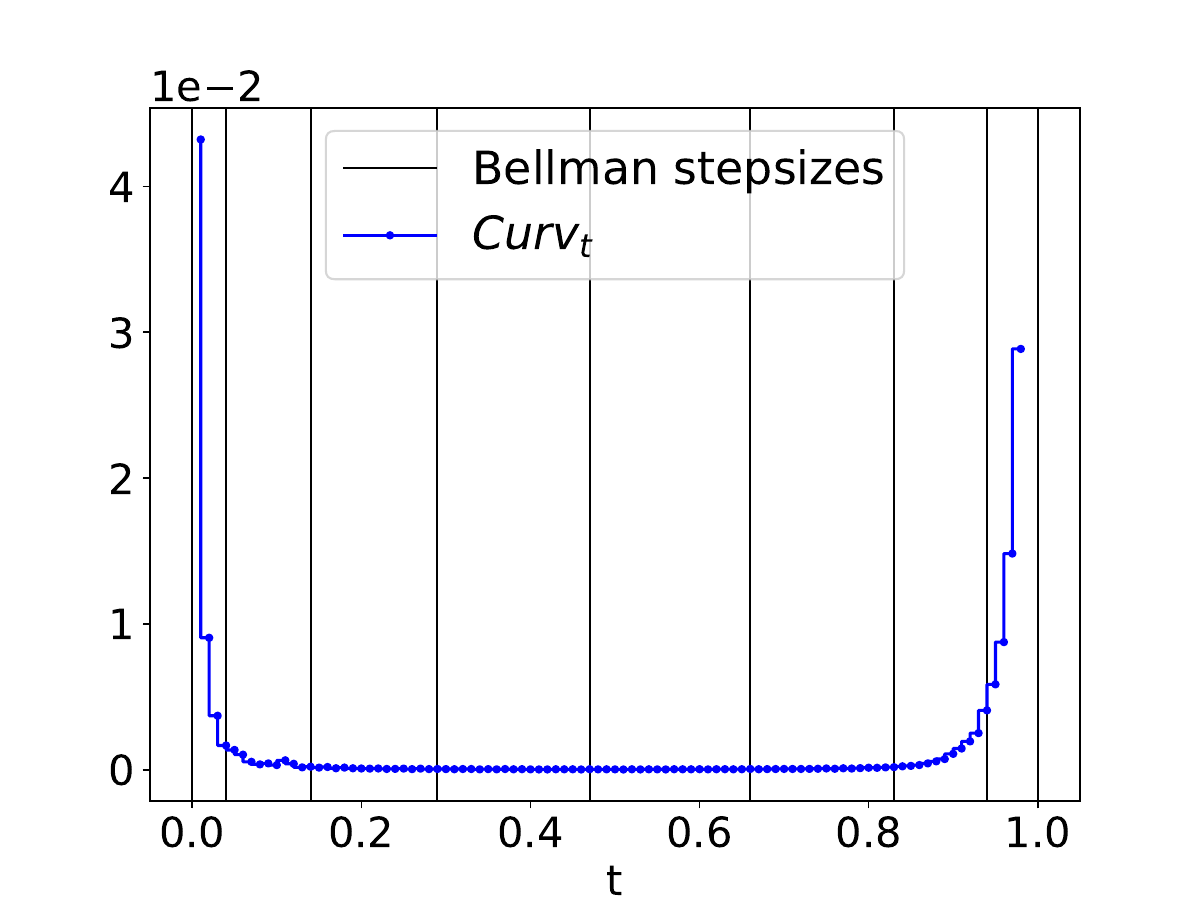}}
    \caption{Curvature measurement (blue curve) and Bellman optimal timestamps (black horizontal line) for different datasets. We observe that the timestamps are denser at regions with higher estimated curvature.}
    \label{fig: curv}
\end{figure}

Subsequently, we plot $\widehat{\mathrm{Curv}}_t$ and investigate whether the Bellman stepsizes coincide with the straightness of the curve, as demonstrated in Figure~\ref{fig: curv}. The curvature of all pretrained rectified models is significant at timestamps close to zero (near the space of noises) and one (near the space of real images). This observation indicates a variant stepsize trajectory, notably at initial and final timestamps. It can be intuitively explained that additional steps are required to match the high-curvature region accurately. Consequently, the outcome of our proposal is sensible, as it correctly identifies the high curvature levels at the beginning and end of the sampling trajectories.

\subsection{Finding Optimal Stepsizes is a Light-weight process}

In this section, we elaborate further on the efficiency of our framework. This efficiency arises because we only need to pass \textbf{a single batch} of noise through the forward process to obtain the values for each intermediate timestamp. Subsequently, we calculate the local truncation error for any two timestamps. These local truncation errors have in total $K^{\max}\times(K^{\max}-1)/2$, and they can be efficiently stored without requiring large memory. The dynamic programming involved in this process is also time-efficient, as elaborated in the Appendix~\ref{sec:dynamic programing}. For added credibility, we provide a running time of the entire stepsizes calculation process with NFE = 10, detailing the running time of each component across all our datasets in Table~\ref{tab:running_time}. All running times are for the Nvidia A5000, an old-generation GPU launched in April 2021. The whole process takes around 115 seconds to complete for the 256x256 datasets.

\begin{table}[H]
\caption{Running time (seconds) for different datasets}
\label{tab:running_time}
\centering
\small
\begin{tabular}{l|c|c|c|c|c}
\toprule
& CIFAR-10 & CelebA-HQ & LSUN-Church & LSUN-Bedroom & AFHQ-Cat \\
\midrule
One batch forward & 8.5977 & 47.7143 & 47.4024 & 48.4324 & 45.5489 \\
Local truncation errors & 10.066 & 67.9143 & 67.5273 & 65.9875 & 66.7532 \\
Dynamic programming & 0.0134 & 0.0138 & 0.0138 & 0.0139 & 0.0138 \\
The whole process & 18.6771 & 115.6424 & 114.9435 & 114.4338 & 112.3159 \\
\bottomrule
\end{tabular}
\end{table}

\begin{table}[H]
\caption{Bellman Optimal Stepsizes for $K=2$ and $K = 4$. The total sum of stepsizes equals one.}
\label{tab:bos24}
\centering
\small
\begin{tabular}{c|c|c}
\toprule
& $K=2$ & $K=4$   \\
\midrule
CIFAR-10 & $[0.28,0.72]$ & $[0.12,0.22,0.31,0.35]$  \\
CelebA-HQ & $[0.13,0.87]$ & $[0.05,0.17,0.32,0.46]$ \\
LSUN-Church & $[0.20,0.80]$ & $[0.08,0.21,0.34,0.37]$ \\
\bottomrule
\end{tabular}
\end{table}

\begin{table}[H]
\caption{Bellman Optimal Stepsizes for $K=6$ and $K = 8$.  The total sum of stepsizes equals one.}
\label{tab:}
\centering
\small
\begin{tabular}{c|c|c}
\toprule
& $K=6$ & $K=8$  \\
\midrule
CIFAR-10 & $[0.09,0.13,0.17,0.21,0.22,0.18]$ & $[0.07,0.09,0.12,0.14,0.17,0.17,0.14,0.10]$ \\
CelebA-HQ & $[0.03,0.08,0.17,0.26,0.28,0.18]$ & $[0.02,0.05,0.11,0.17,0.21,0.20,0.16,0.08]$ \\
LSUN-Church & $[0.06,0.15,0.24,0.26,0.20,0.09]$ & $[0.04,0.09,0.15,0.17,0.18,0.17,0.13,0.07]$ \\
\bottomrule
\end{tabular}
\end{table}

The Bellman steps being far from uniform also means that the probability path of the pretrained models is far from straight, and performing the straightening operation would be beneficial.

We report in this section the Bellman optimal stepsizes obtained in Section~\ref{sec:bellman}. We want to focus on the case $K=8$ to see the common trend of sampling step sizes. Initially, the sampling process takes small step sizes, possibly for structural determination of the images. The stepsizes become larger for the intermediate steps. The last two stepsizes show a decreasing trend: the sampling process takes small stepsizes at the end to refine and potentially make the final output less noisy.

\section{The Transfer of Optimal Stepsizes across Datasets}
To verify the generalization of optimized stepsizes, we transferred the optimized stepsizes from LSUN-Church to the pretrained models on CelebA-HQ and LSUN-Bedroom. The FID scores obtained with 4, 6, and 8 NFEs for CelebA-HQ resulting from this transfer are presented in Table~\ref{tab:celeba_fid_scores}.

\begin{table}[htb]
\centering
\caption{FID scores for CelebA-HQ with different methods and NFEs}
\label{tab:celeba_fid_scores}
\begin{tabular}{l|c|c|c}
\toprule
Method & 4 NFEs & 6 NFEs & 8 NFEs \\
\midrule
Uniform Euler & 158.95 & 127.01 & 109.42 \\
Bellman Euler & 92.03 & 72.54 & 49.80 \\
Bellman-transfer & 132.04 & 100.68 & 72.88 \\
\bottomrule
\end{tabular}
\end{table}

Uniform Euler uses uniform step sizes, Bellman Euler uses optimal stepsizes for CelebA-HQ, while Bellman-transfer uses the stepsizes taken from LSUN-Church. Bellman Euler is still the optimal method. However, what is important here is that Bellman-transfer is better than Uniform Euler. This hints that there is a certain degree of transferability of the step sizes.\footnote{This is an empirical claim; we do not impose any theoretical claim.} In the empirical realm, we can attribute this transferability to a comparable curvature pattern exhibited by pretrained rectified models, as discussed in Section~\ref{sec:explain_steps}.

Table~\ref{tab:lsun_bedroom_fid_scores} is for the LSUN-Bedroom dataset. We observe the same trend here, empirically confirming that the stepsizes have a certain degree of transferability. Nevertheless, optimizing the stepsizes using Bellman Euler would still obtain the best performance.

\begin{table}[htb]
\centering
\caption{FID scores for LSUN-Bedroom with different methods and NFEs}
\label{tab:lsun_bedroom_fid_scores}
\begin{tabular}{l|c|c|c}
\toprule
Method & 4 NFE & 6 NFE & 8 NFE \\
\midrule
Uniform Euler & 84.35 & 39.19 & 32.15 \\
Bellman Euler & 61.60 & 35.35 & 25.80 \\
Bellman-transfer & 70.23 & 38.01 & 29.14 \\
\bottomrule
\end{tabular}
\end{table}

\section{Low-Rank Adaptation for Straightening}
\label{sec:lora}
\begin{figure}[h]
    \centering
    \includegraphics[width=0.5\linewidth]{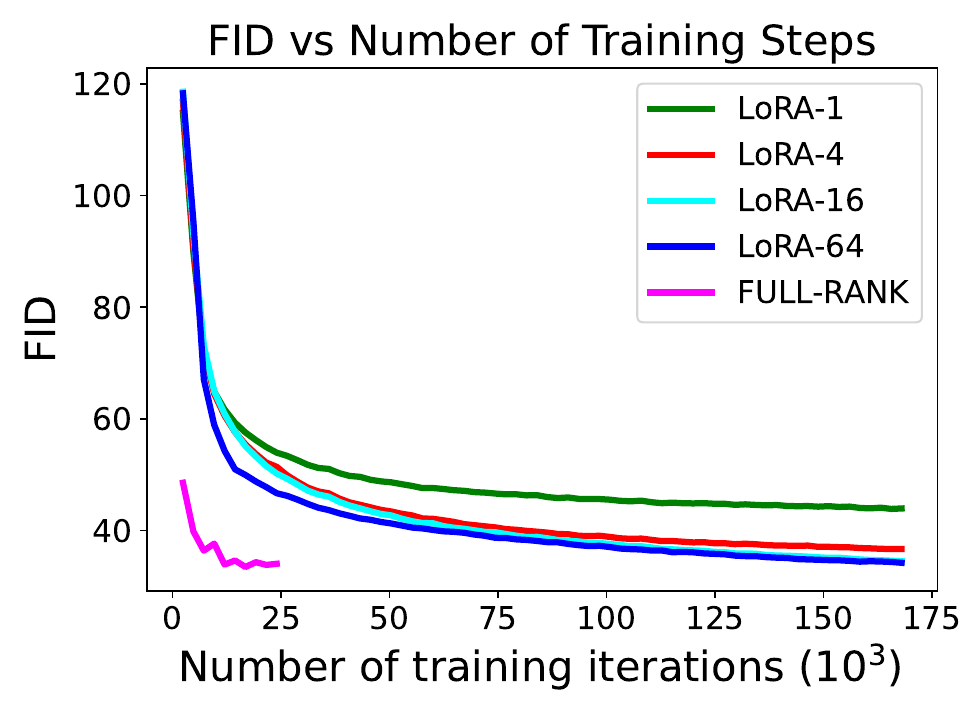}
    \caption{The FID score of straightening procedures along the training iterations on the CelebA-HQ dataset.}
    \label{fig:lora}
\end{figure}

In this section, we expertiment with adding a low-rank adaptation to the linear and convolutional layers of the velocity network $v_\theta$ . For instance, the $t^{\text{th}}$ linear layer represented by an $m \times n$ matrix $W_t$ is adapted to
\[
\widehat W_t = W_t + A_t B_t^{\top},
\]
where $A_t$ is an $m \times r$ matrix, and $B_t$ is an $n \times r$ matrix. The value $r \ll \min\{m, n\}$ represents the rank of the adaptation. This adaptation is similarly applied to convolutional layers, with a slight adjustment: a convolutional layer is first reshaped into a two-dimensional matrix before incorporating the adaptation term. We keep all original parameters of the models fixed and only update the $A_t$ and $B_t$ matrices during the straightening process. We have four versions of straightening named LoRA-$r$, where $r$ is chosen from the set $\{1, 4, 16, 64\}$. Their FID scores on the CelebA-HQ dataset over training iterations are plotted in Figure~\ref{fig:lora}, while the number of trainable parameters for each method is presented in Table~\ref{tab:lora_param}. The FULL-RANK variant described in Section~\ref{sec:straighten} is the method that finetuning \textit{all} parameters of the original model. It is noticeable that versions LoRA-4, LoRA-16, and LoRA-64 can almost match the FID score of the FULL-RANK version, which is 33.86. Specifically, at 175,000 training iterations, LoRA-4, LoRA-16, and LoRA-64 achieve FID scores of 36.70, 34.52, and 34.20 respectively. Furthermore, LoRA-4 only finetunes 2\% of the parameters of the original model but still achieves competitive results compared to the FULL-RANK version.  

\begin{table}[htb]
\centering
\caption{The number of trainable parameters of straightening procedures compared to the full-rank straightening on the CelebA-HQ dataset}
\label{tab:lora_param}
\begin{tabular}{l|c|c|c}
\toprule
Method & Number of Trainable parameters & Percentage on the FULL-RANK version ($\%$) \\
\midrule
LoRA-1 & 330,562 & 0.5 \\
LoRA-4 & 1,322,248 & 2.02 \\
LoRA-16 & 5,288,992 & 8.07 \\
LoRA-64 & 21,155,968 & 32.26 \\
FULL-RANK & 65,574,549 & 100 \\
\bottomrule
\end{tabular}
\end{table}

\section{Additional Qualitative Results}

\begin{figure}[H]
    \centering
    \includegraphics[width=\linewidth]{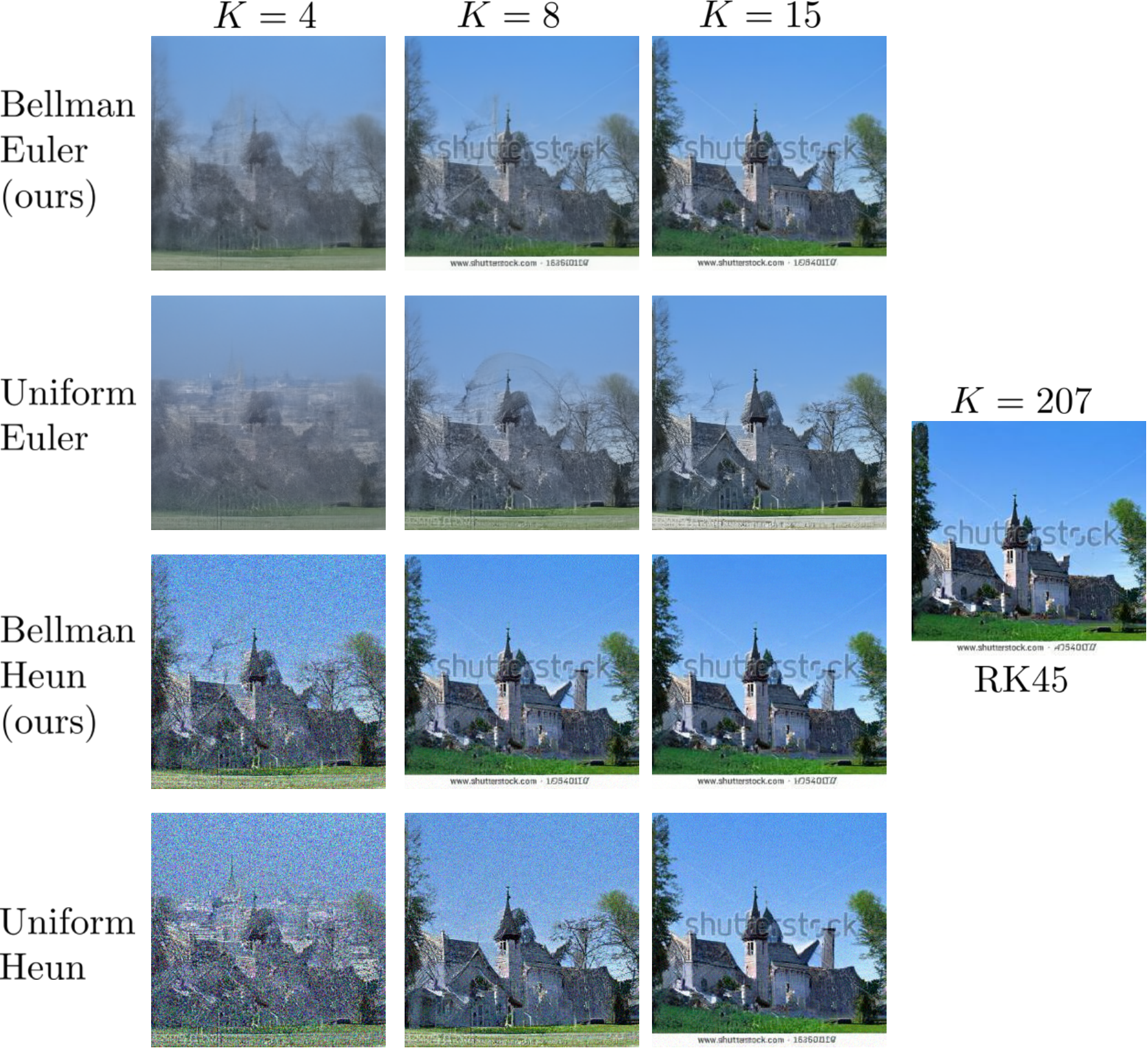}
    \caption{Comparison between images generated from an identical noise with different sampling methods and the number of stepsizes on the LSUN-Church dataset.}
    \label{fig:qual_church}
\end{figure}

\newpage

\begin{figure}[h]
    \centering
    \includegraphics[width=\linewidth]{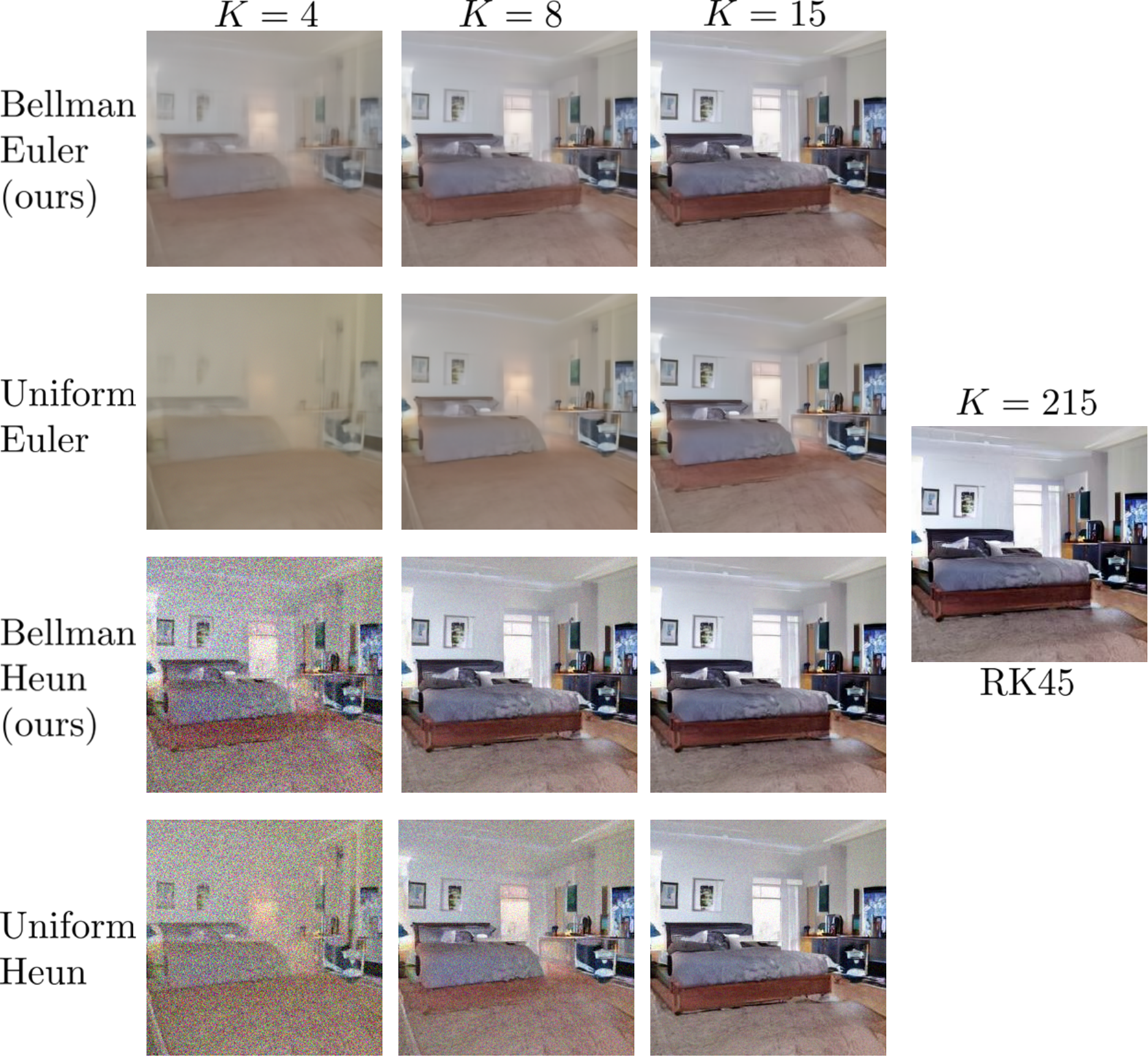}
    \caption{Comparison between images generated from an identical noise with different sampling methods and the number of stepsizes on the LSUN-Bedroom dataset.}
    \label{fig:qual_bed}
\end{figure}
\begin{comment}
\begin{figure}[H]
    \centering
    \includegraphics[width=\linewidth]{retrain_church_2nfe.png}
    \caption{BOSS shows superior performance in qualitative results against other finetuning methods including Uniform-Reflow and Distill-k Reflow with 2 NFE}
    \label{fig:finetune_comp}
\end{figure}
\end{comment}

\begin{figure}
    \centering
    \subfloat[BOSS (NFE = 6, FID = 17.3)]{\includegraphics[width=0.49\linewidth]{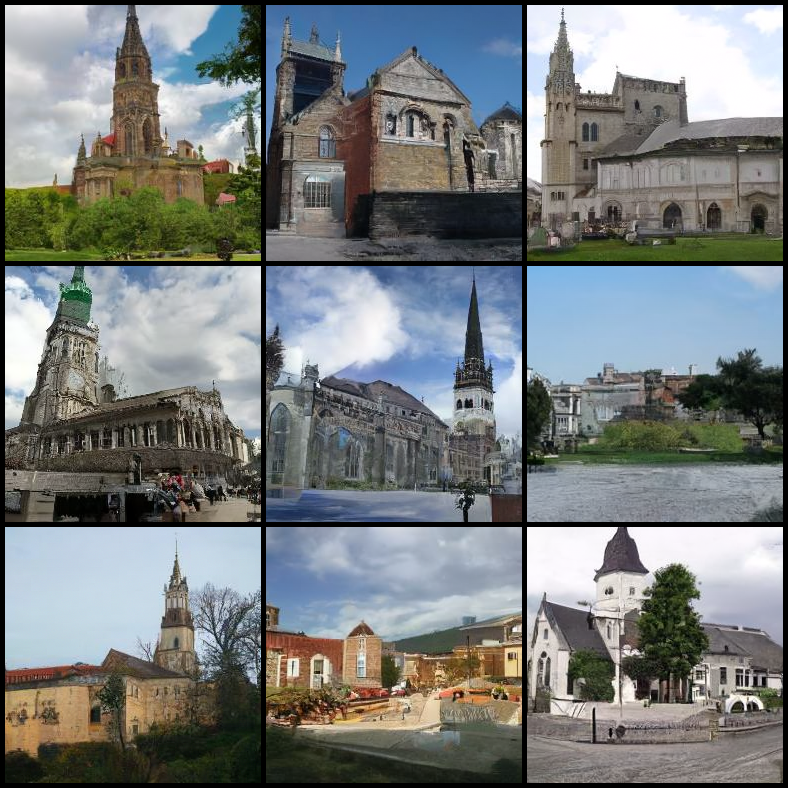}}\hfill
    \subfloat[RK45 (NFE = 203.4, FID = 11.4)]{\includegraphics[width=0.49\linewidth]{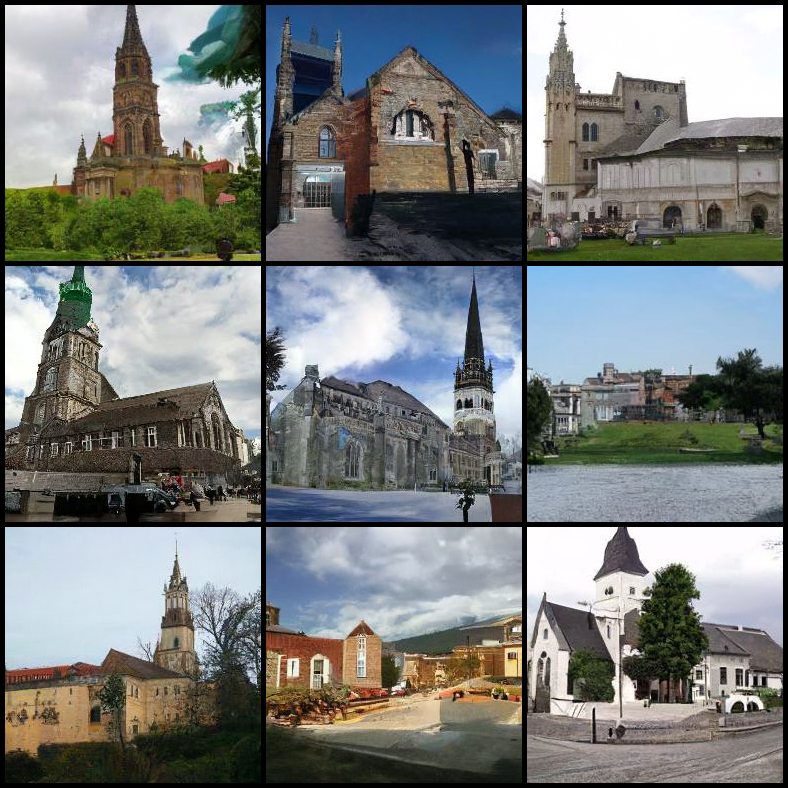}}
    \caption{Comparative qualitative outcomes of BOSS with NFE = 6. The image on the right showcases the generated images referenced by RK45.}
    %\label{fig:three_images}
\end{figure}

\begin{figure}[H]
    \centering
    \includegraphics[width=0.95\linewidth]{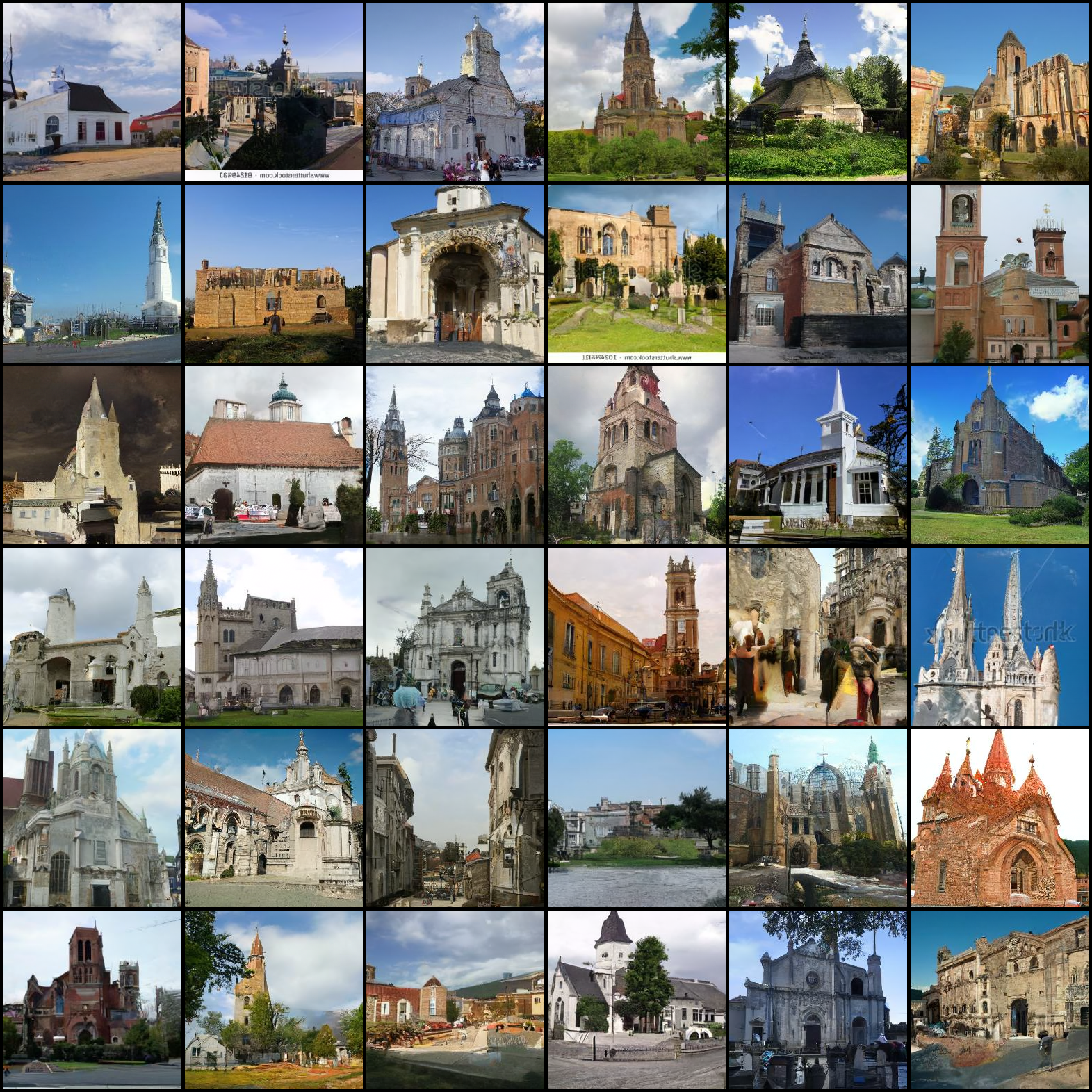}
    \caption{Uncurated images generated from the model finetuned by BOSS (NFE = 10, FID=13.89)}
    \label{fig:redress_10}
\end{figure}

\begin{figure}
    \subfloat[Uniform stepsizes with 4 NFE]{\includegraphics[width=0.49\linewidth]{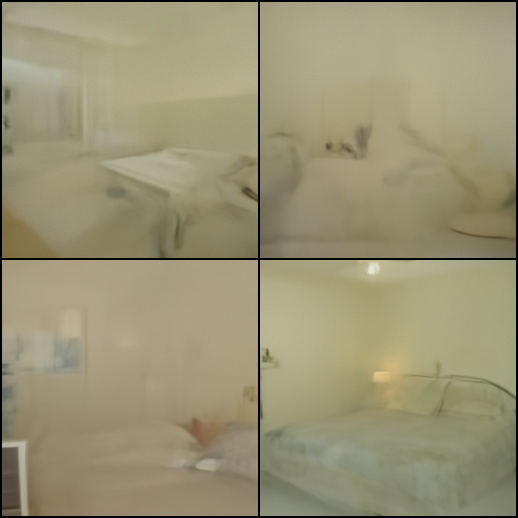}}\hfill
    \subfloat[Bellman optimal stepsizes with 4 NFE]{\includegraphics[width=0.49\linewidth]{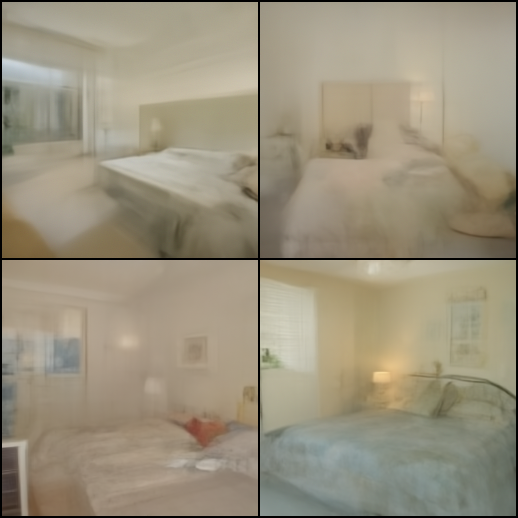}}\hfill
    \subfloat[BOSS with 2 NFE]{\includegraphics[width=0.49\linewidth]{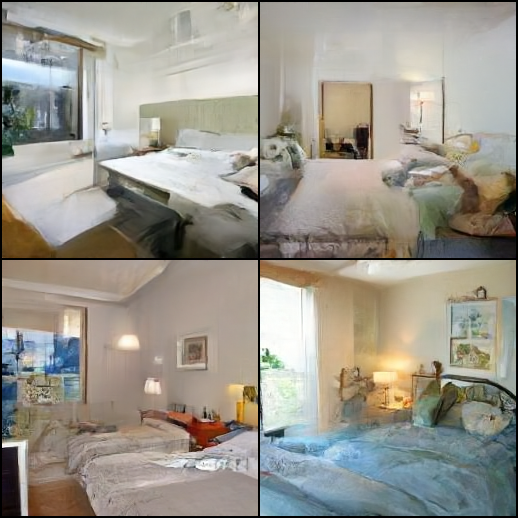}}\hfill
    \subfloat[BOSS with 4 NFE]{\includegraphics[width=0.49\linewidth]{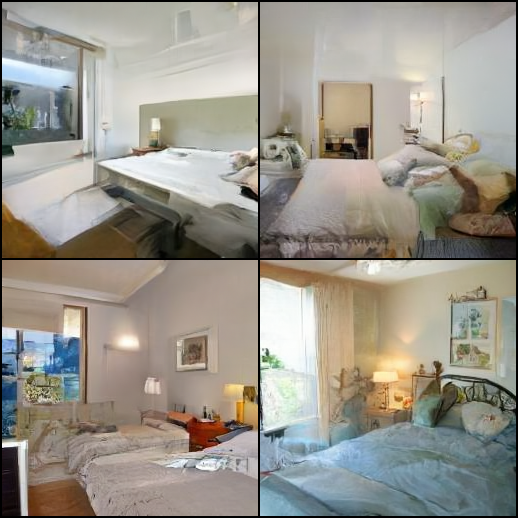}}\hfill
    \subfloat[BOSS with 10 NFE]{\includegraphics[width=0.49\linewidth]{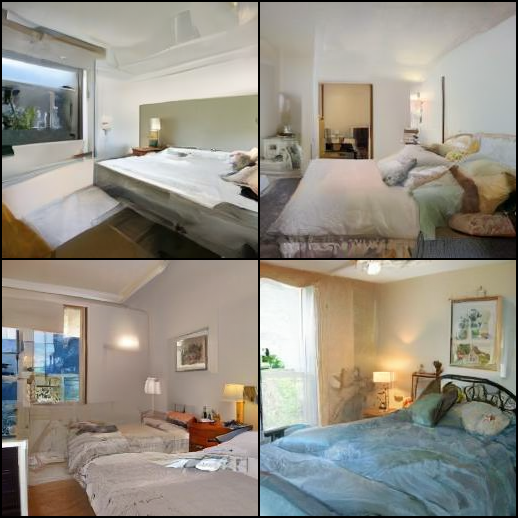}}\hfill
    \subfloat[RK45 with 211.2 NFE]{\includegraphics[width=0.49\linewidth]{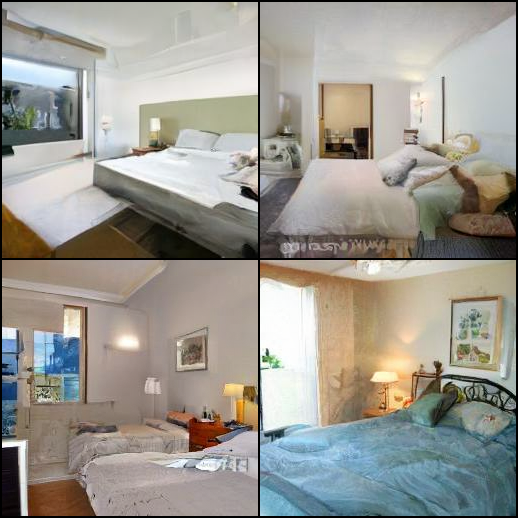}}

    \caption{Comparative qualitative outcomes of BOSS with different NFEs on the LSUN-Bedroom dataset.}
    %\label{fig:three_images}
\end{figure}

\begin{figure}
    \
    \subfloat[Uniform stepsizes with 4 NFE]{\includegraphics[width=0.49\linewidth]{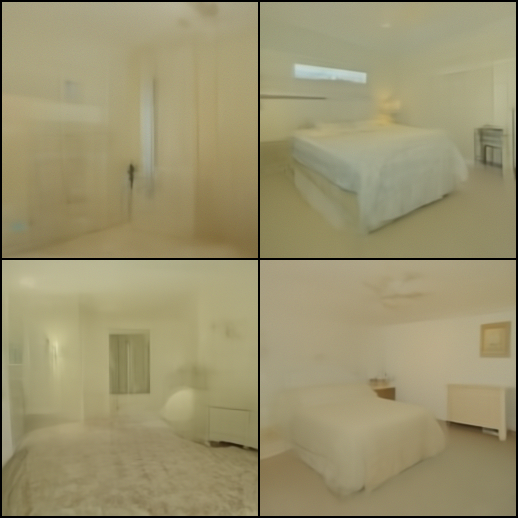}}\hfill
    \subfloat[k-step reflow with 4 NFE]{\includegraphics[width=0.49\linewidth]{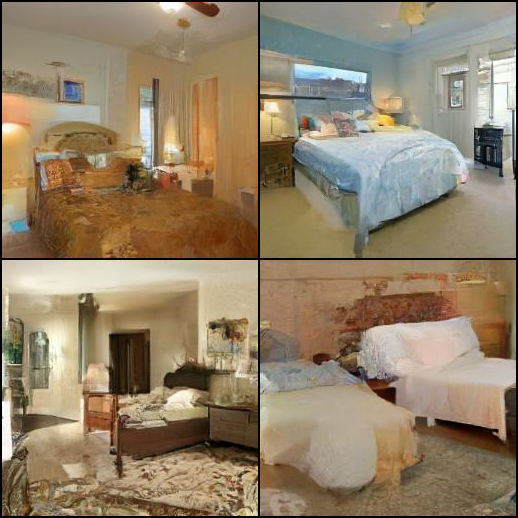}}\hfill
    \subfloat[Bellman optimal sizes with 4 NFE]{\includegraphics[width=0.49\linewidth]{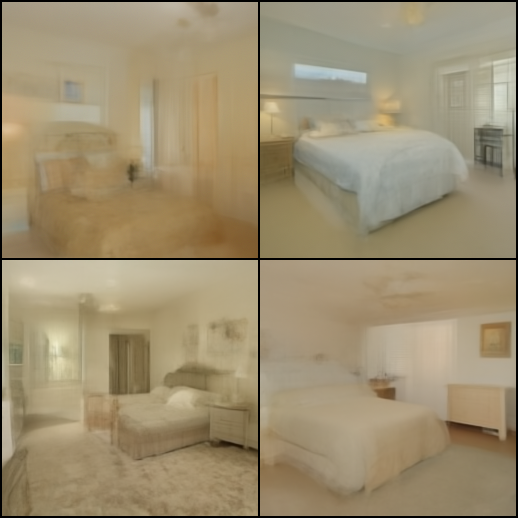}}\hfill
    \subfloat[Uniform Reflow with 4 NFE]{\includegraphics[width=0.49\linewidth]{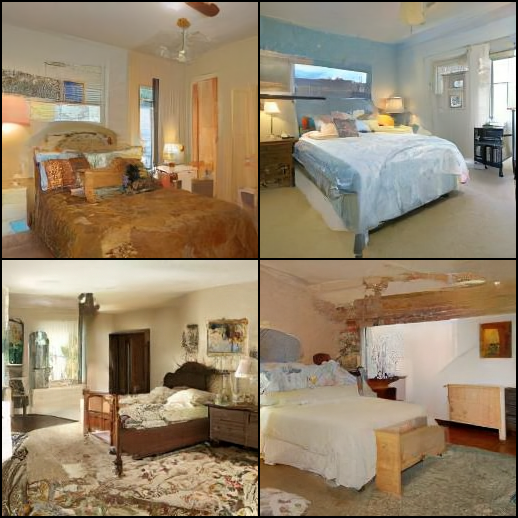}}\hfill
    \subfloat[RK45]{\includegraphics[width=0.49\linewidth]{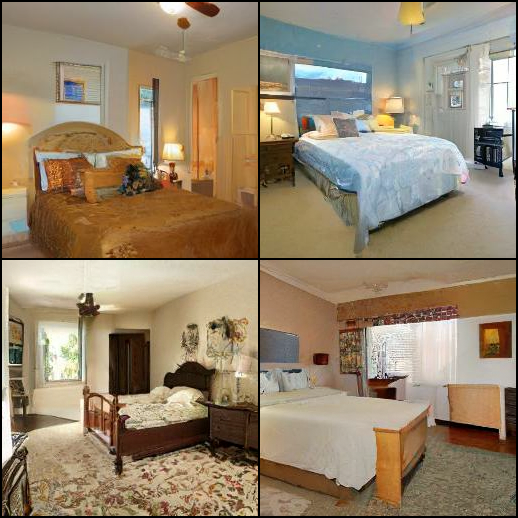}}\hfill
    \subfloat[BOSS with 4 NFE]{\includegraphics[width=0.49\linewidth]{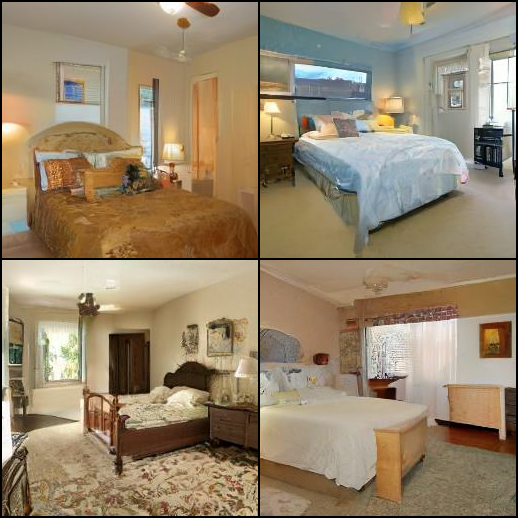}}

    \caption{Samples from LSUN-Bedroom. All corresponding samples use the same initial noise.}
    %\label{fig:three_images}
\end{figure}

\end{document}